\documentclass[sn-mathphys-num]{sn-jnl}

\usepackage{graphicx}%
\usepackage{multirow}%
\usepackage{amsmath,amssymb,amsfonts}%
\usepackage{amsthm}%
\usepackage{mathrsfs}%
\usepackage{dsfont}
\usepackage[title]{appendix}%
\usepackage{textcomp}%
\usepackage{manyfoot}%
\usepackage{booktabs}%
\usepackage{algorithm}%
\usepackage{algorithmicx}%
\usepackage{algpseudocode}%
\usepackage{listings}%
\usepackage{siunitx}
\usepackage[dvipsnames,table,xcdraw]{xcolor}
\usepackage{array}
\usepackage[dvipsnames]{xcolor} % for more color names
\usepackage{subcaption}
\usepackage{comment}

\usepackage{xpatch}
\makeatletter
\xpretocmd{\@maketitle}
{\renewcommand\thefootnote{\hskip1.0ex\@fnsymbol\c@footnote}}
{}{}
\usepackage[symbol*]{footmisc}

\usepackage{graphicx}   
\usepackage{subcaption} 
\usepackage{overpic}
\begin{document}

\title{Physics-consistent machine learning: output projection onto physical manifolds}

\author[1]{\fnm{Matilde} \sur{Valente}}

\author[2]{\fnm{Tiago} \sur{C. Dias}}

\author[1]{\fnm{Vasco} \sur{Guerra}}

\author[3]{\fnm{Rodrigo} \sur{Ventura}}

\affil[1]{\orgdiv{Instituto de Plasmas e Fusão Nuclear}, \orgname{Instituto Superior Técnico, University of Lisbon}, \orgaddress{\street{Av. Rovisco Pais 1}, \city{Lisbon}, \country{Portugal}}}

\affil[2]{\orgdiv{Department of Electrical Engineering}, \orgname{University of Michigan}, \city{Ann Arbor}, \country{USA}}

\affil[3]{\orgdiv{Instituto de Sistemas e Robotica}, \orgname{Instituto Superior Técnico, University of Lisbon}, \orgaddress{\street{Av. Rovisco Pais 1}, \city{Lisbon},  \country{Portugal}}}

\abstract{
%%%%%%%%%
% version, with fine tuning
Data-driven machine learning models often require extensive datasets, which can be costly or inaccessible, and their predictions may fail to comply with established physical laws. Current approaches for incorporating physical priors mitigate these issues by penalizing deviations from known physical laws, as in physics-informed neural networks, or by designing architectures that automatically satisfy specific invariants. However, penalization approaches do not guarantee compliance with physical constraints for unseen inputs, and invariant-based methods lack flexibility and generality.
We propose a novel \textit{physics-consistent} machine learning method that directly enforces compliance with physical principles by projecting model outputs onto the manifold defined by these laws. This procedure ensures that predictions inherently adhere to the chosen physical constraints, improving reliability and interpretability.
Our method is demonstrated on two systems: a spring-mass system and a low-temperature reactive plasma. Compared to purely data-driven models, our approach significantly reduces errors in physical law compliance, enhances predictive accuracy of physical quantities, and outperforms alternatives when working with simpler models or limited datasets.
The proposed projection-based technique is versatile and can function independently or in conjunction with existing physics-informed neural networks, offering a powerful, general, and scalable solution for developing fast and reliable surrogate models of complex physical systems, particularly in resource-constrained scenarios.
}

\keywords{
Physics-informed machine learning, 
Physics-consistent machine learning,
Neural networks, 
Projection method, 
Surrogate models, 
Low-temperature plasma
}

\maketitle

\section{Introduction}\label{sec1} % cercal de 680 palavras

The numerical simulation of physical models is prevalent in science and engineering. These models mathematically represent a physical system, typically by partial differential equations (PDEs) or a set of coupled ordinary differential equations (ODEs). Their development aims at capturing the essential physics of the system, predict its behavior, clarify the principles underlying key observations, and guide experiment design, process optimization and scientific discovery.
Computer simulations explicitly solve the differential equations that result from model development.
However, they often require solving complex systems of equations and become computationally expensive.
Data-driven models have emerged as a promising complement to direct simulations, due to their ability to handle intricate nonlinear input-output relationships. %and construct fast surrogate models.

Machine learning methods are increasingly being used to construct surrogate models for complex physical systems, in disciplines as varied as materials science \cite{Nyshadham2019, Schmidt2019, Jiang2021}, fluid dynamics \cite{Pache2022, Dakane2024, Zhang2025} and low-temperature plasmas \cite{Bonzanini2023, Liu2024, Zhao2024}. 
The reduced computational cost of surrogate models enables real-time control, as quick predictions can be made without execution of the original model. However, these surrogates often require large datasets, which may not be available due to the high cost of data acquisition in practical applications, while their predictive power degrades in the presence of noisy, sparse or dynamic data \cite{Diaw2021}. Moreover, being solely dependent on the data provided during the model training, the predictions may fail to comply with known physical laws.

The introduction of physics-informed neural networks (PINNs) by Raissi et al. \cite{Raissi2019} blends the causality and extrapolation capabilities of physics-based models with the speed, flexibility, and high-dimensional capabilities of Neural Networks (NNs). 
By incorporating physical priors described by differential equations 
into the NN's loss function,
PINNs proved to be effective in addressing a variety of practical engineering and scientific challenges \cite{intro_pinns_flows, intro_pinns_materials, Huang2023, Moschou2023, intro_pinns_fusion}.
Still, as the physical constraints are introduced directly into the NN during training, this approach does not guarantee that the outputs for unseen inputs will satisfy the physical laws after the training process \cite{Kashinath2021, Karniadakis2021, Cuomo2022, Seo2024}. If some properties of the solutions are known, such as \textit{e.g.} energy conservation, it is possible to encode them in the network architecture \cite{Bolton2019, Jin2020, Karniadakis2021, Beucler2021, Tong2021}. However, the need to design specialized network architectures for each system and specific set of constraints makes it difficult to attain a general formulation enforcing adherence to the physical laws. Therefore, although PINNs are currently used with success to solve PDEs, fractional equations, integral-differential equations, and stochastic PDEs \cite{Cuomo2022}, approaches that introduce general but robust physical laws into machine learning (ML) models remain limited.

In this paper, we present a novel approach for \textit{physics-consistent} machine learning, motivated by the current limitations in both purely data-driven methods and physics-informed models. 
Specifically, we focus on studying the effect of projecting the output of an artificial NN onto the manifold defined by a set of chosen physical constraints of the system. 
In this way, our method leverages fundamental physical principles, such as energy or charge conservation, to correct \textit{a posteriori} the predictions of an ML model. Consequently, this method ensures physically consistent predictions and improved accuracy, provided the outputs are reasonably close to the target value.

The projection operation is formulated as a constrained optimization problem,
\begin{equation}
\begin{array}{c}
\underset{p}{\text{minimize}} \quad \|p - f(x;\Theta)\|^2_W \\[10pt]
\text{s.t.} \quad g(x,p) = 0
\end{array}
\label{eq_projection_constraint_problem}
\end{equation}
where the ML parametric model is defined as $y=f(x; \Theta)$ with $x$ the input, $y$ the output, and $\Theta$ is the model parameter vector, 
the physical laws are defined by the constraint vector \(g(x, y) = 0\), 
and $W$ is a symmetric positive definite weighting matrix, i.e., $\|v\|^2_W = v^T W v$, being essentially the metric of the output space. 

The projection method constitutes a highly flexible tool to develop fast and reliable surrogate models of complex physical systems that comply with an arbitrary set of physical laws. 
It enables the use of simpler ML models and/or smaller datasets, while maintaining the same level of predictive accuracy and requiring smaller computational times. Notice that the approach is not restricted to NNs but can also be applied with generality to other ML models, such as Support Vector Regression, Decision Trees, or even simple linear regression models.

We apply the projection method and study its performance in two different physical systems, used as case studies. 
In the first one, we build an artificial NN model to predict the time evolution of a spring-mass system given an arbitrary initial condition. 
The predictions are then corrected by projecting the output on the manifold defined by energy conservation.
The second system is a highly complex and non-linear low-pressure oxygen reactive plasma, created by a DC glow discharge. This system was recently modeled and simulated in \cite{bib_intro_13} and provides an ideal testbed for the proposed approach. 
In this case, we develop an artificial NN  surrogate model according to the reaction mechanism proposed in \cite{bib_intro_13}, and project the output onto the manifold defined by charge conservation, plasma quasi-neutrality and constant operating pressure.

\section{Results and discussion}\label{sec2}

\subsection{Framework}\label{subsec2_1}

\begin{figure}[h!]
\centering
\includegraphics[width=1\textwidth]{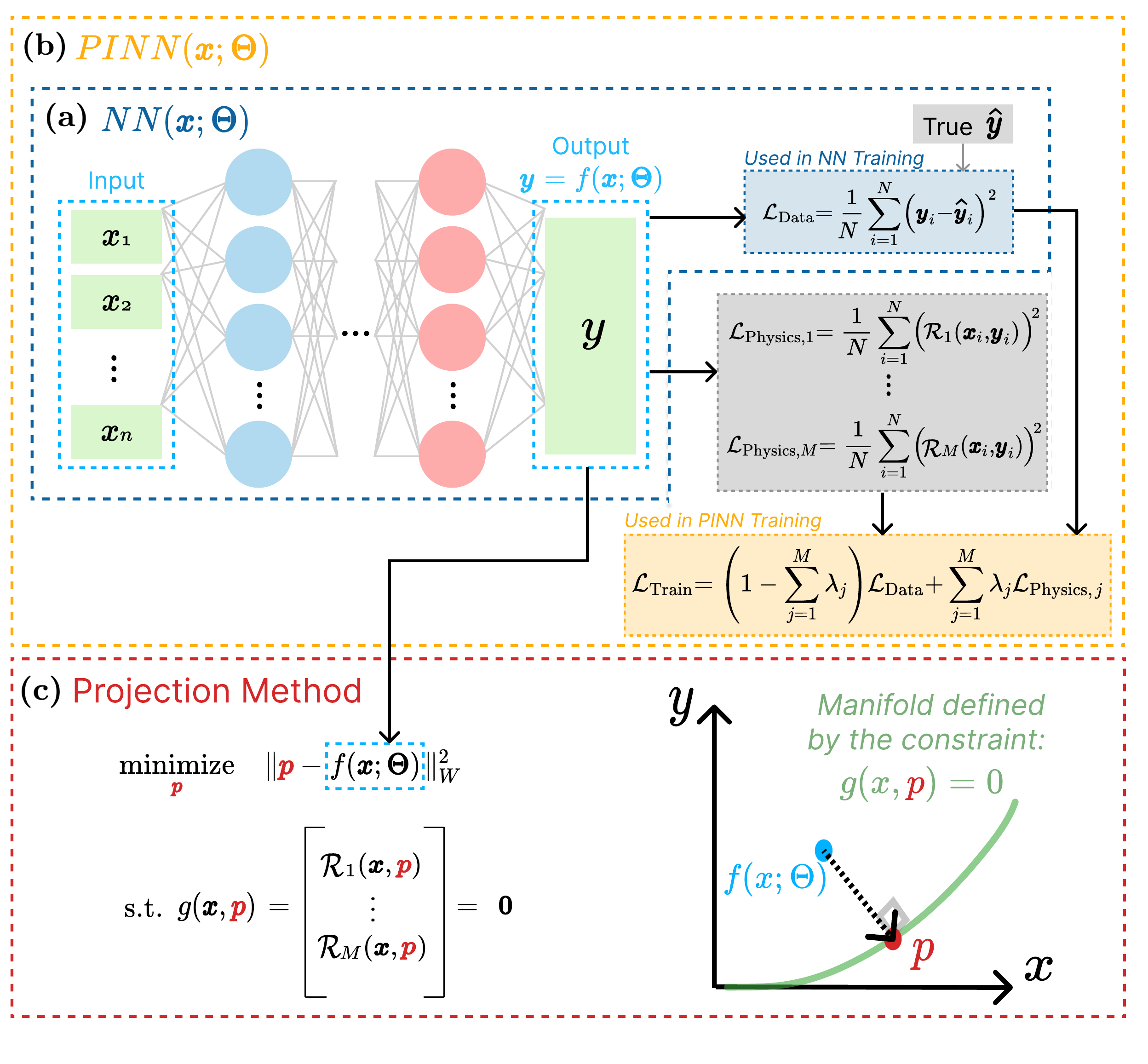}
\caption{\textbf{Schematic overview of the approach in this study.} \textbf{a.} \textcolor{MidnightBlue}{Artificial NN data-driven model} \( y = f(x; \Theta) \), where
\( x \) represents the input vector, \( y \) is the predicted output vector, and \( \Theta \) denotes the model parameter vector.
\textbf{b.} \textcolor{Apricot}{Loss-based PINN model} with a regularization term in the loss function, where \(\hat{y}_i\) represents the target output vector, \(y_i\) represents the predicted output vector, \( \mathcal{L}_{Data}\) represents the data loss term between the PINN's predicted outputs and the target data, \(\mathcal{R}_j\) represents the residual associated with the \(j^{th}\) physical law, \( \mathcal{L}_{Physics, j}\) represents the physics loss term regarding the \(j^{th}\) physical law imposed to the system, \( \mathcal{L}_{Train}\) represents the total training loss function, which is a weighted sum balancing data fitting and adherence to physical constraints, and \(\lambda_j\) represents the weight factor given to the \(j^{th}\) physical law, with \(\lambda = \sum_{j=1}^{M} \lambda_j \leq 1\).
\textbf{c.} Formulation and visualization of the \textcolor{BrickRed}{projection operation method} as a constraint optimization problem, with the projection of the output \(y\) of the model onto the manifold defined by the constraint vector \(g(x, y) = 0\), where \(W\) is a symmetric positive weight matrix.}\label{three_model_flowchart}
\end{figure}

A schematic overview of our approach is illustrated in Fig. \ref{three_model_flowchart}.
Both systems under study are described by a set of coupled ODEs, and their
numerical solution is used as the ground truth. 
We then compare the predictions of four surrogate models: an artificial NN data-driven model, a loss-based PINN (where the physics priors are included in the loss function), and the projection method applied to the predictions of each of the two former models. We compare the quality of the prediction of the model outputs before and after applying the projection, analyze the robustness across varying model complexity and training, and discuss the physical insight into the underlying physics. Fig. \ref{three_model_flowchart}c also makes clear the intuition behind the projection method.

\subsection{Case study 1: spring-mass system}\label{subsec2_2}

This example highlights the ability of the projection method to handle sequential predictions and how it corrects the trajectories as they gradually deviate from the target values.

\subsubsection{System description}\label{subsec2_2_1}

The system consists of two masses, \( m_1 \) and \( m_2 \), connected in series by two springs with spring constants \( k_1 \) and \( k_2 \), and natural lengths \( L_1 \) and \( L_2 \), respectively. 
The first spring is connected to a fixed wall at one end and to mass \( m_1 \) at the other, while the second spring connects the two masses.
The masses are restricted to moving along the $x$-axis and there is no friction.
The positions of \( m_1 \) and \( m_2 \) along the $x$-axis are denoted by \( x_1 \) and \( x_2 \), and their velocities by \( v_1 \) and \( v_2 \), respectively. The forces exerted by the springs on the masses are determined by the displacements from their equilibrium positions, following Hooke's law.

The differential equations of motion are given by
\begin{equation}
\begin{cases}
    m_1 {\ddot{x}}_1 = -k_1 (x_1 - L_1) + k_2 (x_2 - x_1 - L_2) \\
    m_2 {\ddot{x}}_2 = -k_2 (x_2 - x_1 - L_2)
\end{cases}\ ,
\label{PDEs_spring_mass_system}
\end{equation}
where \( \ddot{x}_1 \) and \( \ddot{x}_2 \) denote the accelerations of masses \( m_1 \) and \( m_2 \). The time-evolution of the system is determined from the solution of these equations, with the initial conditions 
\begin{equation}\label{initialConditions}
\begin{cases}
    x_1(0) = x_{0,1}, \quad x_2(0) = x_{0,2} \\
    v_1(0) = v_{0,1}, \quad v_2(0) = v_{0,2}
\end{cases}\ .
\end{equation}
Since there is no friction, the mechanical energy, $E$, defined by the sum of the kinetic energy %(\(K\)) 
of the masses and the elastic potential energy, %(\(U\)), 
is conserved.
Hereafter we consider $m_1=$ 1 kg, $m_2=$ 1 kg, $k_1=$ 5 N/m, $k_2=$ 2 N/m, $L_1=$ 0.5 m and $L_2=$ 0.5 m.

\subsubsection{Data generation}\label{subsec2_2_2}

The artificial NN data-driven model takes as input a \textit{state vector} characterizing the state of the system at the current instant $t$ (\textit{i.e.}, defining the positions and velocities of both masses at time $t$), and predicts the state vector at time $t+\Delta t$, for fixed $\Delta t$. 
This procedure enables the recursive determination of the complete trajectory from a given initial condition, with time-resolution $\Delta t=50$~ms, as a sequence of transitions between states.
We used an energy threshold \(E_{max}=\) 5 J to create a set of allowed states for the system, corresponding to the range of positions and velocities the masses can have with $E<E_{max}$. 
Eq. (\ref{PDEs_spring_mass_system}) was solved using the classical fourth-order Runge-Kutta method (RK4), with arbitrary initial conditions (\ref{initialConditions}) ensuring that $E<E_{max}$.

Except otherwise noted, the dataset consists of \(N=\) 100000 arbitrarily generated input state vectors within the energy threshold and the corresponding next states, determined by performing a single Runge-Kutta computation up to time $\Delta t$ (not to be confused with the Runge-Kutta time-step).
To reduce the impact of the difference in feature magnitude on the model and make the training process more stable, we applied the min-max normalization. Consequently, all features were scaled to the range \([-1,1]\). 

\subsubsection{Trajectory prediction}\label{subsec2_2_3}

\begin{figure}[h!tbp]
    \centering
    
    \begin{subfigure}[b]{1\textwidth}
        \centering
        \begin{overpic}[width=\textwidth]{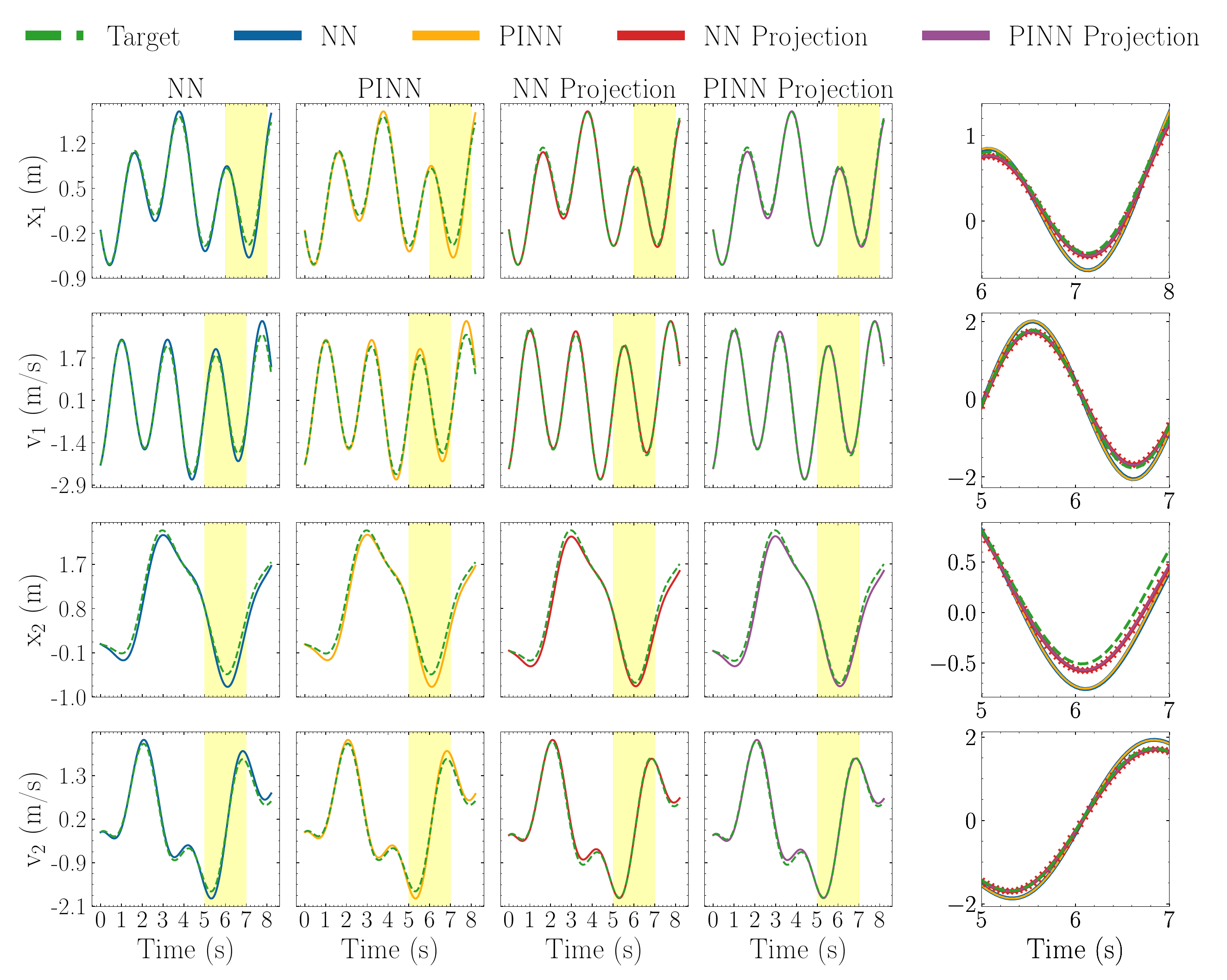}
            \put(1,82){\textbf{(a)}} 
        \end{overpic}
        \label{fig:b}
    \end{subfigure}

    \begin{subfigure}[b]{0.37\textwidth}
        \centering
        \begin{overpic}[width=\textwidth]{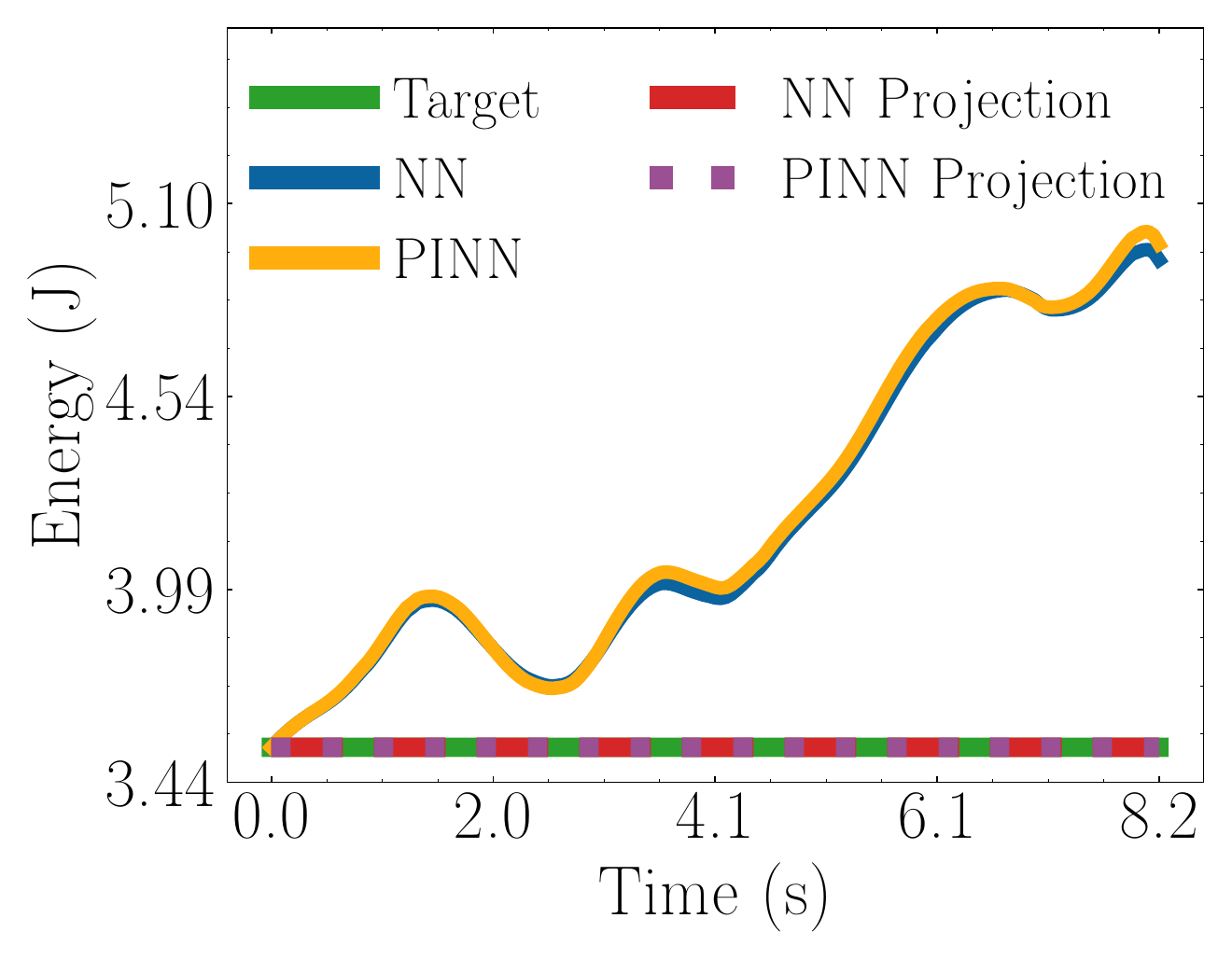}
            \put(-1,83){\textbf{(b)}}
        \end{overpic}
        \label{fig:c}
    \end{subfigure}
    \hfill
    \begin{subfigure}[b]{0.60\textwidth}
        \centering
        \begin{overpic}[width=\textwidth]{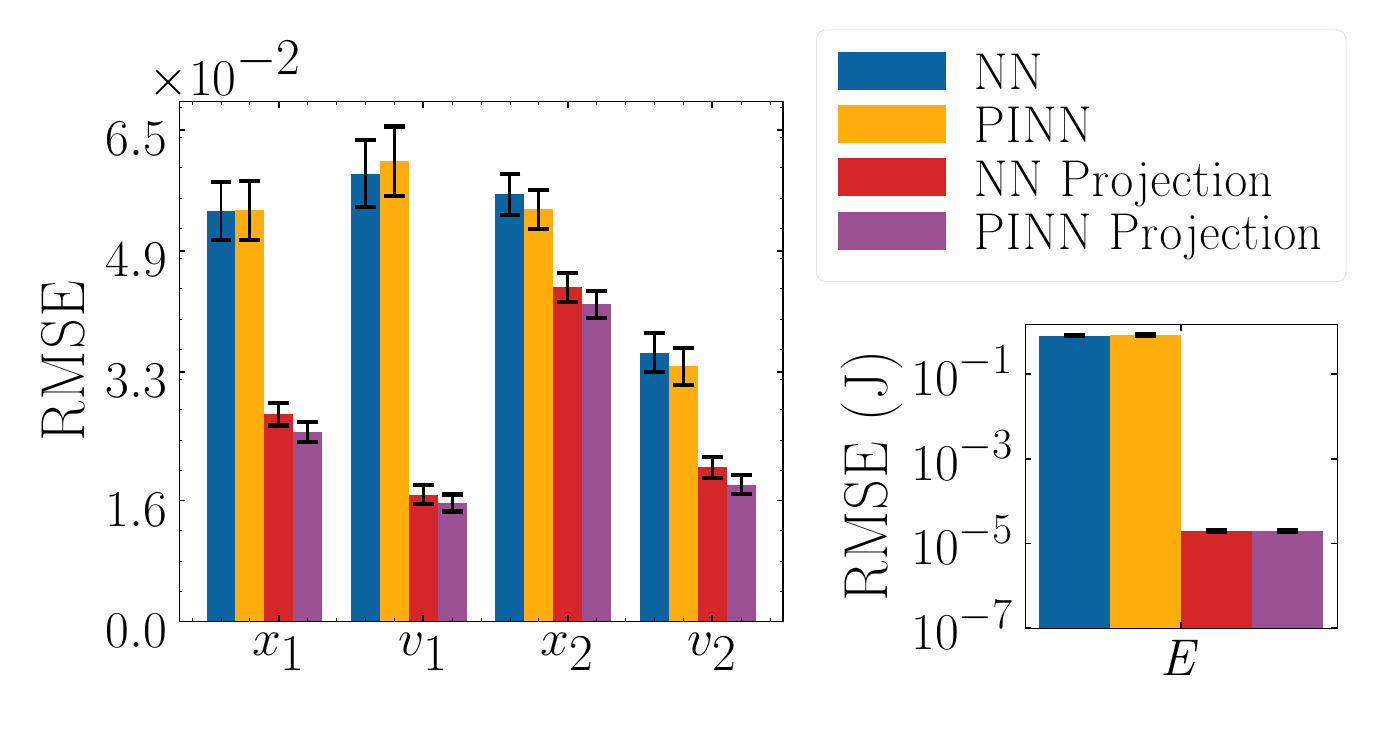}
            \put(0,50){\textbf{(c)}} 
        \end{overpic}
        \label{fig:d}
    \end{subfigure}
    \caption{\textbf{Comparative analysis of the models' performance for a given initial condition.}  
    \textbf{a.} Predicted and target positions and velocities over time using the four models considered: NN (first column), loss-based PINN (second column), projection method applied to the NN outputs (third column), and projection method applied to the PINN outputs (fourth column).
    The yellow-shaded areas highlight regions where the NN and loss-based PINN deviate from the target, demonstrating the projection method's corrective performance. 
    The right column provides zoomed-in views of the yellow-shaded regions.
    \textbf{b.} Predicted and target total energy over time using the four models. 
    \textbf{c.} Bar plot comparing the RMSE for positions, velocities, and energy for the four models considered. The RMSE values are calculated using normalized positions and velocities, while the energy RMSE is computed in Joules, with the initial energy as the target.}
    \label{fig:Fig_2}
\end{figure}

The trajectory of the system is defined by the prediction of its state along 165 sequential time steps, i.e., 8.25 seconds.
The general physical law considered in the loss-based PINN and in the projection method is 
energy conservation, which translates into the residual \(\mathcal{R}_1 = E(t) - E(0)\).
The results of the four models on the four state-variables and energy conservation are presented in Fig. ~\ref{fig:Fig_2}a-b, respectively, for the arbitrary initial condition $x_{0,1} = \SI{-0.16}{m}$, $x_{0,2} = \SI{0.09}{m}$, $v_{0,1} = \SI{-2.18}{m/s}$ and $v_{0,2} = \SI{-0.16}{m/s}$.
The projection method applied to the outputs of the NN reduces the root mean squared error (RMSE) between the normalized predictions and normalized target values on the four state variables, by 
49.5\%, 
71.7\%, 
21.7\%, 
42.6\% 
for \(x_1\), \(v_1\), \(x_2\), and \(v_2\), respectively, when compared with the purely data-driven NN model. 

The predictions are not only more accurate, but also physically consistent, showing a reduction in the RMSE for the energy conservation law by more than four orders of magnitude when compared with the NN model, reducing the error from $(7.99 \pm 0.35) \times 10^{-1}$~J for the NN to $(1.96 \pm 0.11) \times 10^{-5}$~J after the projection operation, as shown in Fig. \ref{fig:Fig_2}c.
Additionally, the loss-based PINN shows a RMSE for energy conservation of $( 8.16 \pm 0.36) \times10^{-1}$~J, of the same order of magnitude as the data-driven model alone.

\begin{figure}[h!tbp]
\centering
\includegraphics[width=1\textwidth]{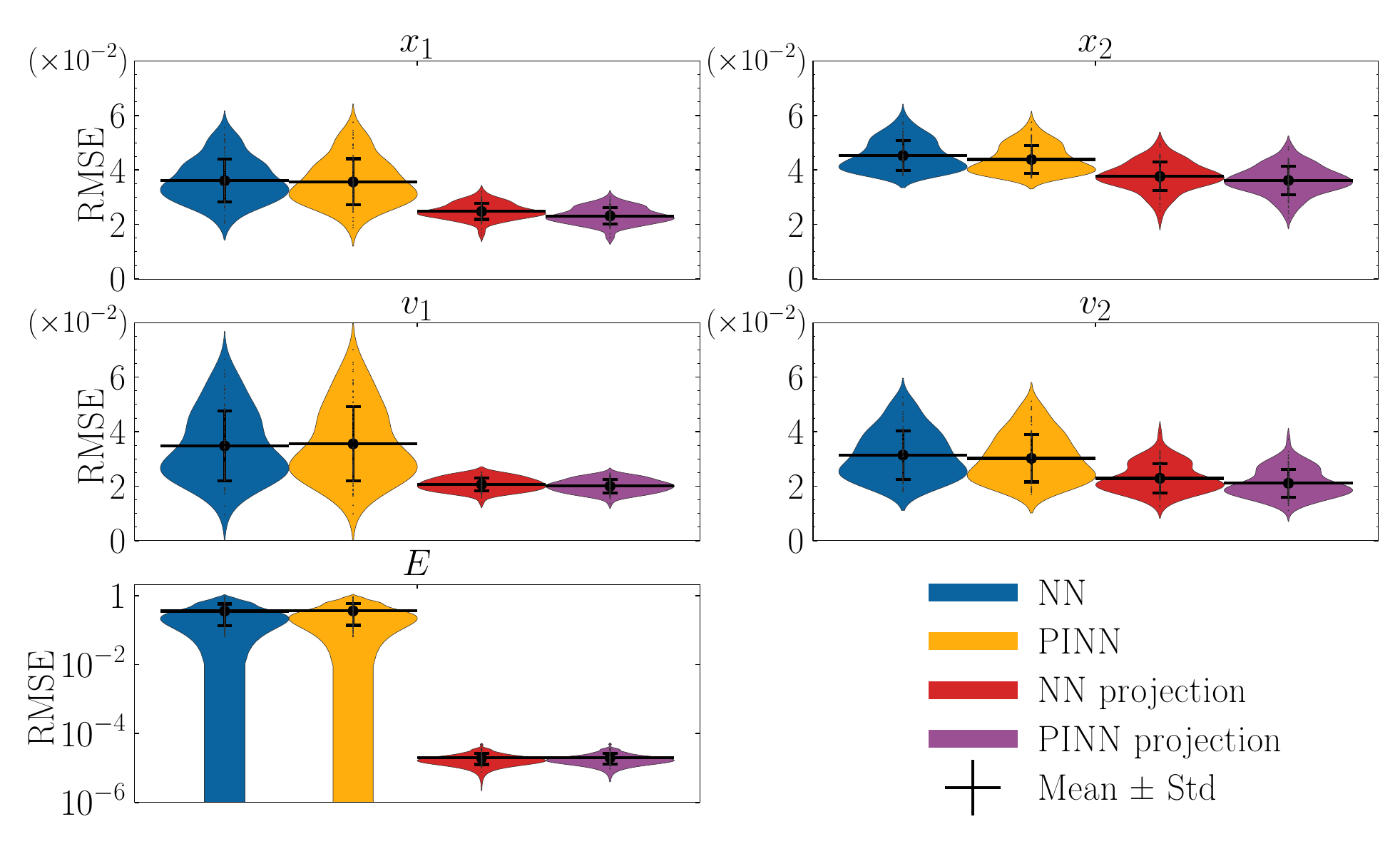}
\caption{\textbf{Comparative analysis of the models' performance for several initial conditions.} Distributions of root mean square error (RMSE) of the 4 models for each normalized state variable and for energy conservation. Each violin plot contains 100 trajectories and each error bar corresponds to the standard deviation of their RMSEs. 
}\label{plot_several_initial_conditions}
\end{figure}

To evaluate the robustness of the projection method, we tested 100 different arbitrary initial conditions from the test set and computed the corresponding trajectories for 10 seconds. 
The performance metrics across state variables and energy conservation for all four models are represented as violin plots in Fig. \ref{plot_several_initial_conditions}. 
Clearly, the projected outputs consistently outperform their non-projected counterparts. 
When comparing the four models, the NN and PINN projections achieve similar RMSEs for all state variables, with the average RMSEs reduced by approximately
33\%, 
42\%,
17\%, and
29\%
for $x_1$, $v_1$, $x_2$ and $v_2$, respectively, compared with the non-projected counterparts.
Moreover, both projection methods demonstrate a remarkable improvement in energy conservation, with a reduction in associated error by more than four orders of magnitude, from approximately
$ 4 \times 10^{-1}$~J to
$ 2 \times 10^{-5}$~J, showcasing the ability of the method to achieve physics-consistent predictions. 
Finally, by analyzing the standard deviations of the RMSEs, which correspond to the error bars presented in Fig. \ref{plot_several_initial_conditions}, we conclude that the projection method also enhances the the prediction stability, as evidenced by consistently smaller standard deviations across most parameters (the only exception is the standard deviation of $x_2$ when comparing the PINN with its projected counterpart, which slightly increases from $0.51 \times 10^{-2}$ to $0.52 \times 10^{-2}$).

Additional information is given in Table \ref{table_improvement_rates_spring_mass_system}, where we indicate the percentage of initial conditions in which the projection method leads to an improvement in the RMSE averaged over the four state variables ($R_{mean}$) and in the RMSE of all of the state variables simultaneously ($R_{all}$).
Both the NN and PINN projections produce identical improvement rates compared to their respective base models, with $\sim 96\%$ of the trajectories showing enhanced $R_{mean}$ performance and $\sim 60\%$ showing improvement in $R_{all}$. 
These results indicate that the projection method consistently provides better predictions than non-projected ones and suggest its benefits are comparable regardless of the underlying base model (NN or loss-based PINN).

\newpage

\begin{table}[!htbp]
    \centering
    \caption{Percentage of arbitrary initial conditions where the mean RMSE across the four state variables ($R_{mean}$) improved, and where the RMSE improved simultaneously for all the four state variables ($R_{all}$), for: projected NN compared with NN and projected PINN compared with PINN.}
    
    \arrayrulecolor{black} % Ensures table lines are black and distinct
    \rowcolors{3}{gray!15}{white} % Alternating row colors
    \setlength{\tabcolsep}{20pt} % Adjust cell padding for better fit
    \renewcommand{\arraystretch}{1.2} % Adjust row height for clearer separation
    
    \begin{tabular}{l|c|c}
        \toprule
        \rowcolor{gray!30} % Darker header color
        \textbf{Rates} & \textbf{NN \(\to\) NN Projection} & \textbf{PINN \(\to\) PINN Projection} \\
        \midrule
        $R_{mean} (\%)$  & 96 & 97 \\ 
        $R_{all} (\%)$   & 60 & 61 \\ 
        \bottomrule
    \end{tabular}

    \label{table_improvement_rates_spring_mass_system}

\end{table}

\newpage

%\enlargethispage{3\baselineskip}
Several aspects of this case study are worth highlighting. 
First, the purely data driven NN and the loss-based PINN yield similar results in the prediction of the four state variables. 
Second, the loss-based PINN does not reduce the error in energy conservation compared to the purely data-driven NN, despite having the corresponding constraint in the loss function.
Third, the projection method consistently provides better predictions than the non-projected ones and its benefits are comparable regardless of the underlying base model (NN or loss-based PINN).
These trends are evident in Fig.~\ref{plot_several_initial_conditions} and by comparing the first and second columns in Table~\ref{table_improvement_rates_spring_mass_system}.
These findings suggest that incorporating very general physical principles, such as energy conservation, directly into the loss function as regularization terms may not provide sufficient physical guidance to the model post-training during testing. 
Somewhat counterintuitively, such approach likely expands rather than constrains the accessible output parameter space, introducing optimization challenges that impair the model's ability to generalize to new observations. 
Consequently, while the data loss gradient pushes the model parameters towards the ground truth output values, the inclusion of a physics loss associated to a general law will likely perturb the gradient direction away from the ground truth.
Finally, the projection enforces energy conservation, reducing the associated error by 4 orders of magnitude.

\subsection{Case study 2: low-temperature reactive plasma}\label{subsec2_3}

This example demonstrates the ability of the projection method to handle complex, high-dimensional and strongly nonlinear systems. 
Additionally, we draw on the underlying knowledge of the plasma system to interpret in physical terms the impact of the projection operation on the model outputs.

\subsubsection{System description}\label{subsubsec2_3_1}

The system under study is an oxygen (O\(_2\)) low-temperature reactive plasma (LTP) created by a continuous DC glow discharge operating at pressures in the range $P\in[0.1,10]$~Torr, %($[13.33, 1333]$~Pa), 
discharge current $I\in[5,50]$~mA, in a long cylindrical tube of radius $R\in[4,20]$~mm. As is typical for low-temperature molecular plasmas, the system exhibits a variety of coupled energy pathways and elementary processes, such as electron impact excitation and de-excitation, gas phase and heterogeneous reactions, dissociation, molecule formation, ionization and charge transfer. A detailed set of reactions and corresponding rate coefficients, validated against benchmark experiments, the so-called reaction mechanism, was recently developed by T.C. Dias et al. \cite{bib_intro_13}, where the experimental data for validation are also given. Herein we consider the kinetic scheme from \cite{bib_intro_13} without vibrational excitation, which accounts for 12 species -- electrons, ground-state molecules O$_2(X)$ and atoms O$(^3P)$, electronically excited states O$_2(a\ ^1\Delta_g$, $b\ ^1\Sigma_{g}^{+}$, Hz) and O($^1D)$, ground-state ozone O$_3$ and vibrationally excited ozone O$_3^\star$, negative ions O$^-$, and positive O$_2^+$ and O$^+$ ions -- and more than 85 elementary processes. 
With the exception of the electron density, $n_e$, the steady-state concentrations of each species $n_s$ are obtained from the solution to the coupled system of ODEs
\begin{equation}
    \frac{dn_s}{dt}=\sum_i\left[\left(a_{si}-b_{si}\right)k_i\prod_j n_j^{b_{ji}}\right]\ ,
\end{equation}    
where $a_{si}$ and $b_{si}$ are the stoichiometric coefficients of species $s$, as they appear on the left- and right-hand  sides of a reaction $i$, respectively. In addition, the average gas temperature $T_g$ and the gas temperature near the tube wall $T_{nw}$ are calculated from the gas thermal balance equation, the reduced electric field $E/N$, where $N$ is the gas density, from the quasi-neutrality condition, the electron drift velocity $v_d$ and temperature $T_e$ as integrals over the electron energy distribution function, and the electron density from the discharge current. Further details are given in \cite{bib_intro_13}. % and in section \ref{subsec4_2}. 

\subsubsection{Data generation}\label{subsubsec2_3_2}

The models include 3 input features ($P$, $I$ and $R$) and the 17 outputs just described.
We generated the datasets with the LisbOn KInetics Chemistry+Boltzmann (LoKI-B+C) simulation tool \cite{bib_intro_3,bib_intro_13}.
%LoKI is a simulation tool for plasma chemistry that couples two main calculation blocks: a Boltzmann solver (LoKI-B) for the two-term electron Boltzmann equation and a Chemical solver (LoKI-C) for the heavy-species kinetics. 
Except otherwise noted, the dataset comprises \(N=\) 1000  uniformly distributed values across the three input features within their specified boundaries.
Similarly to the previous case, we applied the min-max normalization to the features, scaling them to the range \([-1,1]\). Moreover, we applied a log-transformation to the features demonstrating skewness in its distribution. Further details on data generation and preprocessing are given in section \ref{subsec4_3}.

\subsubsection{Prediction of the steady-state plasma properties}\label{subsubsec2_3_3}

The general physical laws constraining the system include: the ideal gas law, relating the gas pressure (input) with the species densities and the gas temperature (outputs); the imposed discharge current (input), expressing electric charge conservation and relating its input value with the tube radius (input), and with the drift velocity and electron density (outputs); and the quasi-neutrality law, relating the electron density (output) with the positive and negative ion densities (outputs). These laws can be represented by the residuals in equations (\ref{eq_ideal_gas_law}--\ref{eq_quasi_neutrality}), respectively,
\begin{equation}
    {\cal R}_1 = P - \sum_i [X_i] k_B T_g
\label{eq_ideal_gas_law}
\end{equation}
\begin{equation}
    {\cal R}_2 = I - e n_e v_d \pi R^2
\label{eq_discharge_current}
\end{equation}

\vspace{0.9em} 

\begin{equation}
    {\cal R}_3 = n_e - \sum_i [X_i^+] + \sum_j [X_j^-]
\label{eq_quasi_neutrality}
\end{equation}

\noindent where %\(P\) is the input pressure, 
$[X_i]$ is the density of species $i$,
\(\sum_i [X_i]\) is the total gas density given by the sum of all the species densities, \(k_B\) is the Boltzmann constant, %\(T_g\) is the gas temperature, \(I\) is the input discharge current, 
\(e\) is the charge of the electron, %\(n_e\) is the output electron density, \(v_d\) is the output drift velocity, \(R\) is the input radius of the reactor, 
\(\sum_i [X_i^+]\) is the sum of the positively charged ion population, and \(\sum_j [X_j^-]\) is the sum of the negatively charged ion population.

\begin{figure}[htbp]
    \centering
    \begin{subfigure}[b]{0.95\textwidth}
        \centering
        \begin{overpic}[width=\textwidth]{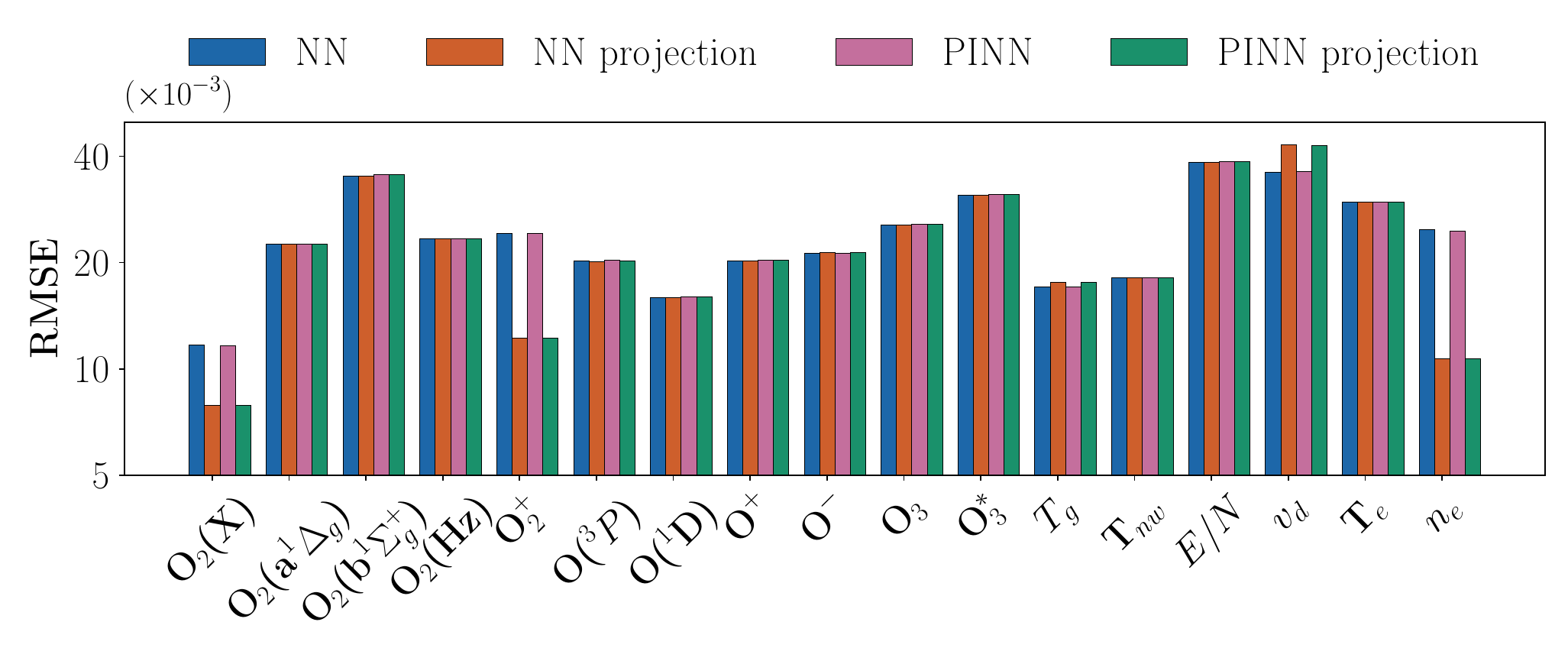}
            \put(-2.5,38){\textbf{(a)}} 
        \end{overpic}
        \label{fig_4a}
    \end{subfigure}
    
    \begin{subfigure}[b]{0.6\textwidth}
        \centering
        \begin{overpic}[width=\textwidth]{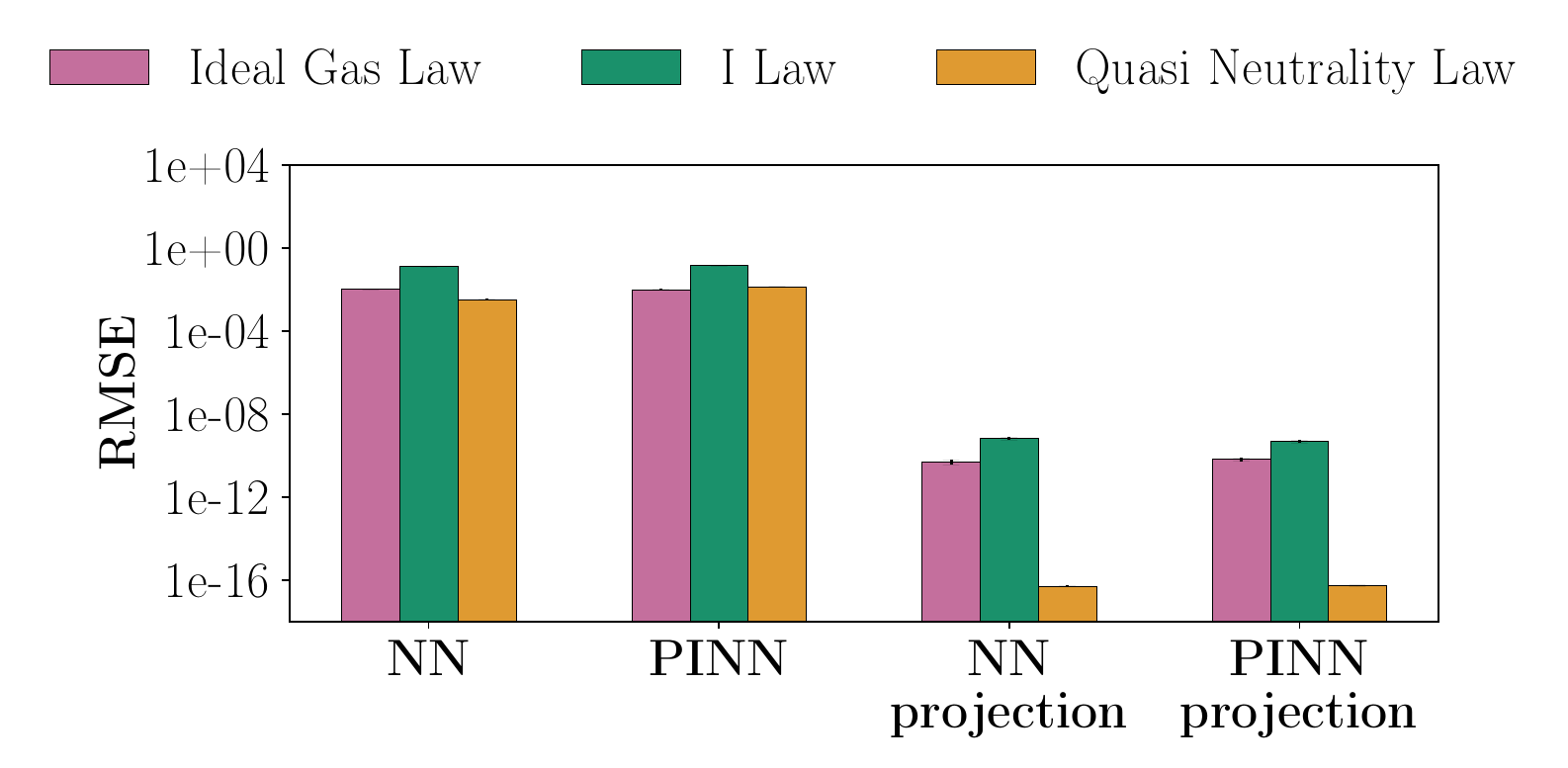}
            \put(-4,50){\textbf{(b)}}
        \end{overpic}
        \label{fig_4b}
    \end{subfigure}
    \begin{subfigure}[b]{0.35\textwidth}
        \centering
        \begin{overpic}[width=\textwidth]{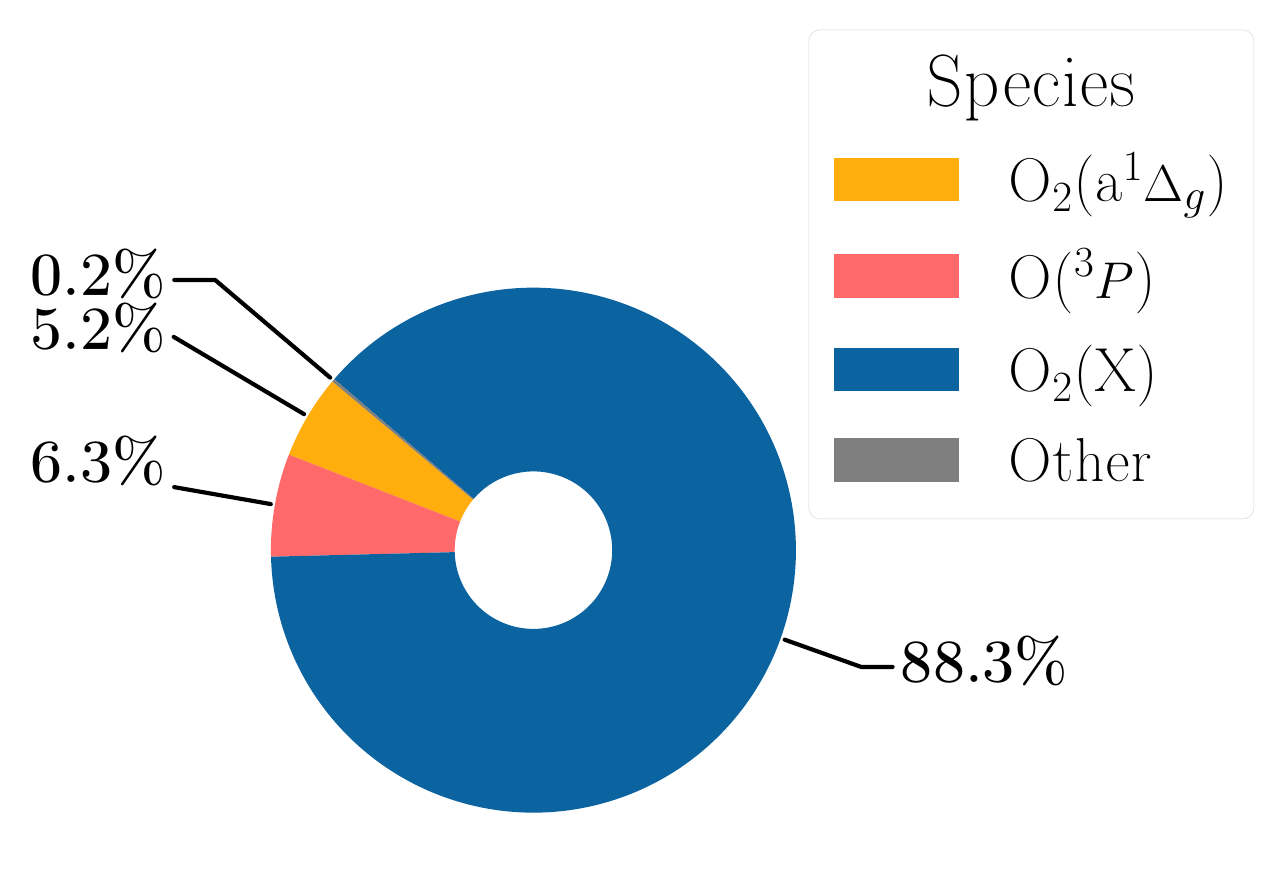}
            \put(-1,80){\textbf{(c)}} 
        \end{overpic}
        \label{fig_4c}
    \end{subfigure}
    
    \begin{subfigure}[b]{1\textwidth}
        \centering
        \begin{overpic}[width=\textwidth]{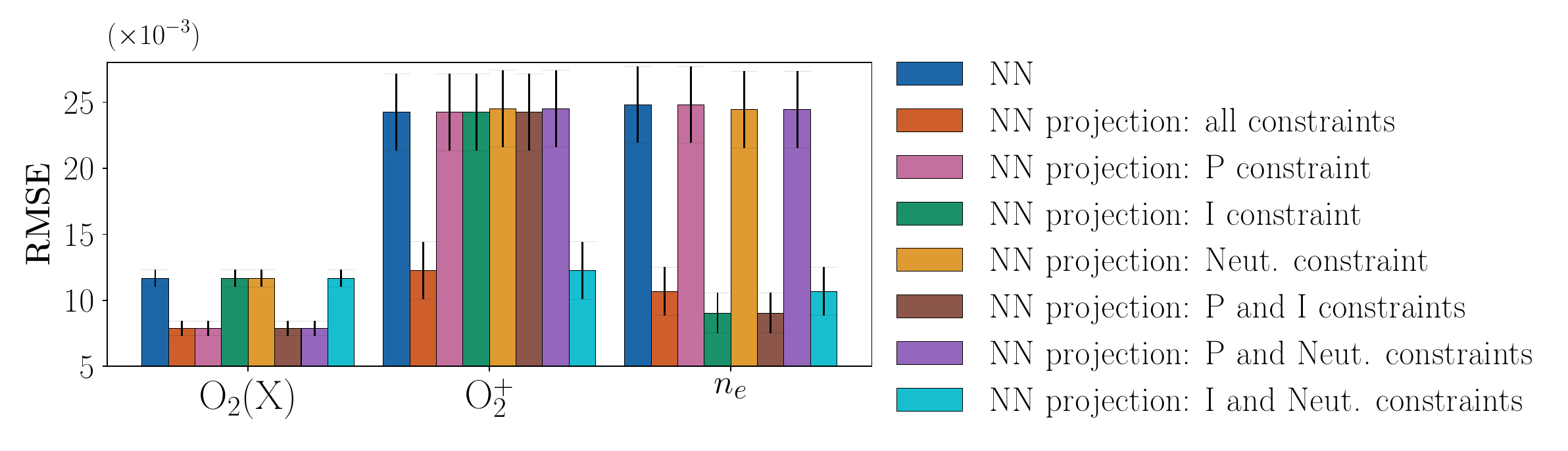}
            \put(-1,28){\textbf{(d)}} 
        \end{overpic}
        \label{fig_4d}
    
    \end{subfigure}

    \caption{\textbf{Steady-state feature estimation in the LTP system.} \textbf{a.} Test set results of the four models (NN, loss-based PINN, projected NN, and projected PINN) when predicting the 17 outputs. %Compared to the other three methods, the NN projection has smaller prediction errors on three outputs. 
    \textbf{b.} Test set results of the four models when evaluating the compliance with physical laws. %The projection method is able to reduce the MAPE in compliance with physical laws by more than eight orders of magnitude. 
    \textbf{c.} Relative density of the main species in the mixture at steady-state. \textbf{d.} Bar plot of the three outputs that improved the most when the projection method was applied to the NN predictions. For each output is presented the RMSE of: the NN, the projection when the three constraints are applied simultaneously, the projection when the constraints are applied individually, and the projection when the constraints are applied in pairs.}
    \label{plots_ltp_analysis}
\end{figure}

The test-set results of the four models on the prediction of the 17 output variables and on the three physical laws are shown in Fig. \ref{plots_ltp_analysis}a-b, respectively, expressed as Root Mean Squared Errors (RMSE). 
Applying the projection method to NN predictions reduces the RMSE of compliance with physical laws by more than 9 orders of magnitude, while it significantly improves the predictive accuracy of three of the output variables: O$_2(X)$, O$_2^+$, and $n_e$. As it was the case in the spring-mass system, imposing very general physical laws is ineffective in guiding the loss-based PINN model, which is not able to reduce the RMSE of compliance with physical laws (Fig. \ref{plots_ltp_analysis}b) nor of the predicted outputs, when compared with the NN model. 

Fig. \ref{plots_ltp_analysis}c-d provide insight into the physical interpretation of the results, by representing the main species in the mixture and examining the effects of each physical constraint individually.
Ground-state O$_2(X)$ molecules constitute 88.3\% of the total mixture and the prediction of their density improves primarily due to the ideal gas law constraint (\ref{eq_ideal_gas_law}), which explicitly accounts for species densities. 
Similarly, the electron density $n_e$ benefits from the imposition of the discharge current constraint (\ref{eq_discharge_current}), as it appears directly in the expression of this law.
Finally, O$_2^+$ constitutes less than 0.0001\% of the gas mixture and, as expected, remains unaffected by the ideal gas law (\ref{eq_ideal_gas_law}). 
It is the dominant positive ion by at least 2 orders of magnitude \cite{bib_intro_13}, making its prediction inherently sensitive to the quasi-neutrality condition (\ref{eq_quasi_neutrality}). However, this constraint alone is insufficient to correct the prediction, as the electron density sets an ``anchor'' for the total ion density. Therefore, the accuracy of the O$_2^+$ density in the projection method is coupled with the prediction of $n_e$, and both the discharge current (\ref{eq_discharge_current}) and quasi-neutrality (\ref{eq_quasi_neutrality}) laws are required to significantly improve the model's performance regarding this output.

\subsection{Robustness of the projection method}\label{subsec2_4}

With sufficient training time and a sufficiently complex architecture, a neural network can produce predictions that closely approximate the target values, reducing (or even eliminating) the need for post-training corrections. 
Conversely, a poorly trained model may yield predictions that deviate significantly from the manifold defined by the physical laws, making the projection operation ineffective or leading to ambiguities associated with multiple possible solutions. 
In this section, we analyze the robustness of the projection method with respect to model complexity, quantified by the number of NN parameters (i.e., weights and biases) and the size of the training dataset. 
The results focus on the low-temperature reactive plasma system described in section \ref{subsec2_3}.

\subsubsection{Ablation study}\label{subsec2_4_1}

We start by comparing the errors before and after applying the projection operation to the NN predictions, considering models with different complexities.
In order to guarantee a comparable analysis between the models, we considered 18 different architectures, each with 2 hidden layers and a number of neurons ranging from 1 to 1000 in the hidden layers. Both hidden layers have the same number of neurons. For each architecture, 1 NN model was trained with a dataset consisting of \(N=\) 1000 data points.

Fig. \ref{fig:Fig_6}a.i shows the RMSE of the predictions of the NN and of the projection method as a function of the number of parameters in the model. These values are calculated by averaging the errors across the 17 outputs.
%The ``RMSE variation rate'' is defined here as the RMSE relative variation after performing the projection and is considered as negative. 
The ``RMSE variation rate'' is defined as the relative change in RMSE after performing the projection and is considered negative if the RMSE decreases.
Fig. \ref{fig:Fig_6}a.ii provides a similar analysis, but for the 3 outputs that improved the most after applying the projection operation, as opposed to the mean of all outputs.
The graphs show that the NN projection consistently achieves lower RMSE than the standard NN. The improvement is particularly pronounced for O$_2(X)$, $\mathrm{O_2^+}$ and $n_e$ predictions, in simpler models with fewer parameters ($\sim 10^2$). 
As the number of parameters increases to $10^5$, both methods reach a performance plateau and the advantage of the projection method becomes marginal, though it still yields slightly better predictions. 
These results suggest that the projection method is particularly beneficial for more constrained, and hence computationally more efficient, NN architectures.

It is instructive to analyze the predictions of pressure-dependent trends for fixed discharge current and tube radius.
Fig. \ref{fig:Fig_6}b.i-iii illustrates how NN architectures of increasing complexity predict $n_e$ as a function of pressure for $I=30$~mA and $R=12$~mm, and benchmarks these predictions to LoKI simulation values. 
In the simplest architecture (Fig. \ref{fig:Fig_6}b.i), the predictions of the NN deviate significantly from the targets in both magnitude and trend (RMSE = $ 2.55 \times 10^{-2}$). Despite this poor initial performance, the projection successfully aligns the predictions with the LoKI targets (RMSE = $0.53 \times 10^{-2}$), albeit with visible lack of smoothness. In the intermediate case (Fig. \ref{fig:Fig_6}b.ii), the projection corrects the noisy NN predictions with remarkable accuracy.
Finally, in the most complex architecture (Fig. \ref{fig:Fig_6}b.iii), the NN starts to align with the target values, indicating that sufficient model complexity can compensate for the absence of explicit physical information, though at the cost of increased computational resources.

\subsubsection{Small samples}\label{subsec2_4_2}

\begin{figure}[htbp]
   \centering
   \begin{subfigure}[t]{0.3\textwidth}
       \centering
       \raisebox{12pt}{
           \begin{overpic}[width=\textwidth]{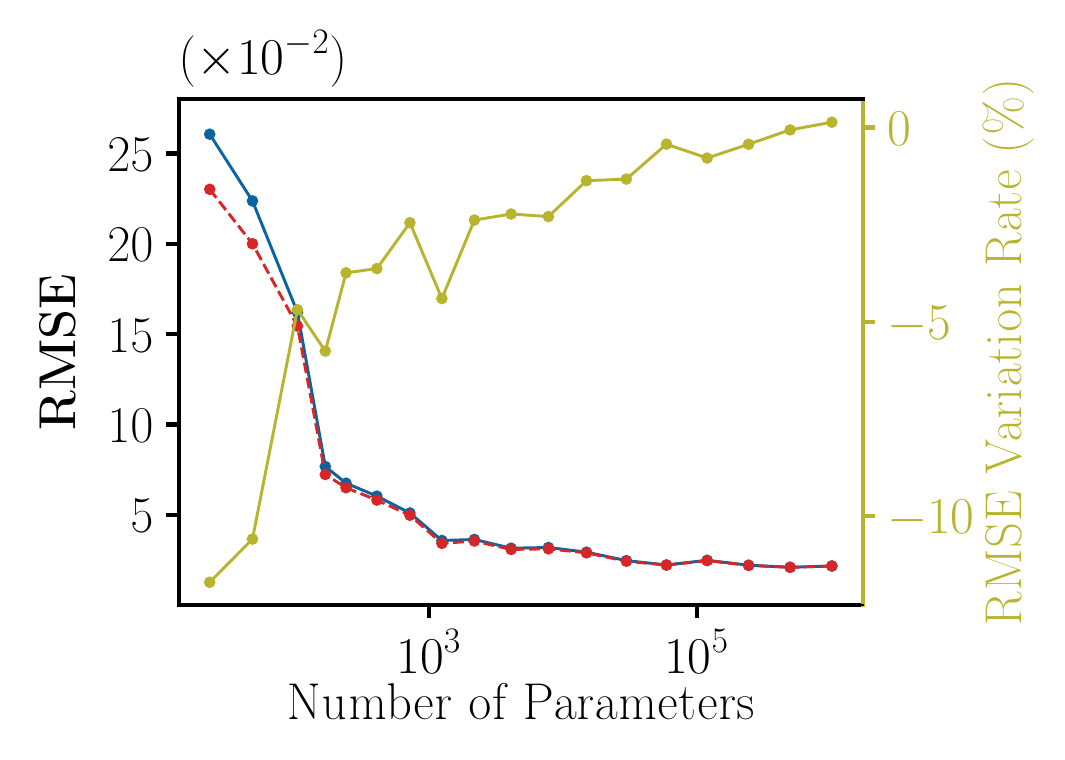}
               \put(2,80){\textbf{(a.i)}}
           \end{overpic}
       }
       \label{fig:4b}
   \end{subfigure}
   \hfill
   \begin{subfigure}[t]{0.68\textwidth}
       \centering
       \begin{overpic}[width=\textwidth]{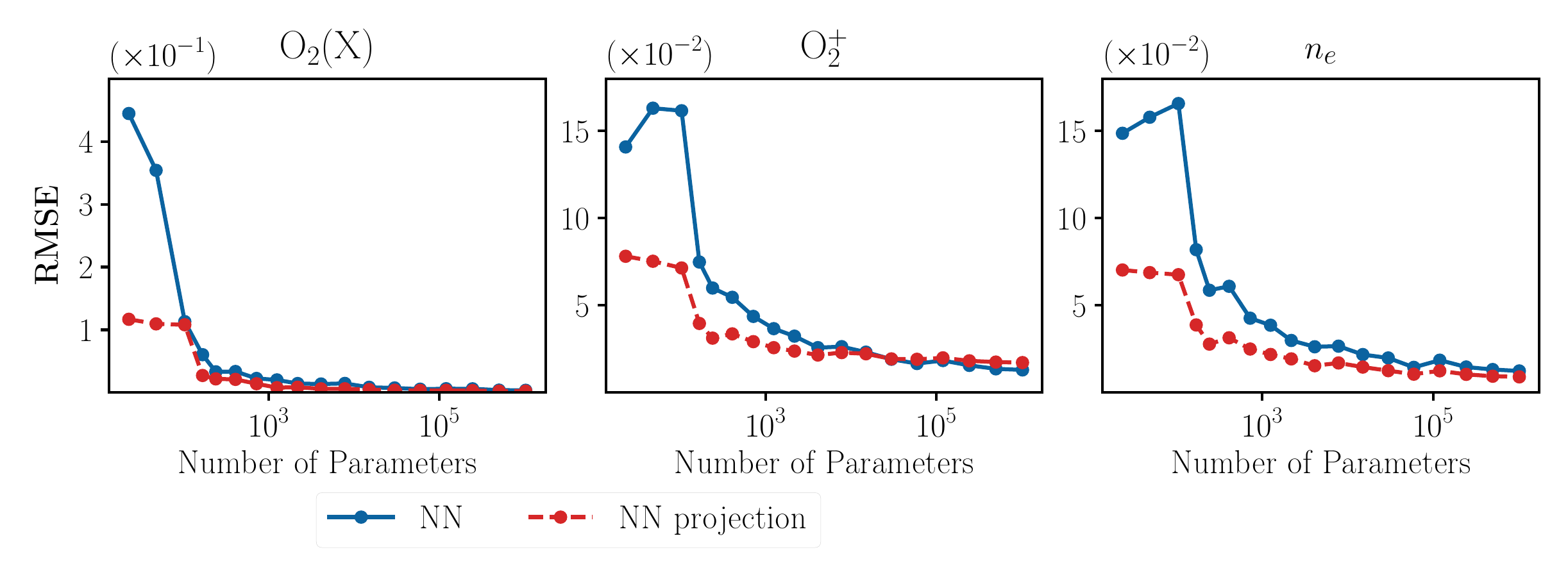}
           \put(2,40){\textbf{(a.ii)}}
       \end{overpic}
       \label{fig:4c}
   \end{subfigure}
   
   \vspace{1em}

    \begin{subfigure}[b]{0.32\textwidth}
        \centering
        \begin{overpic}[width=\textwidth]{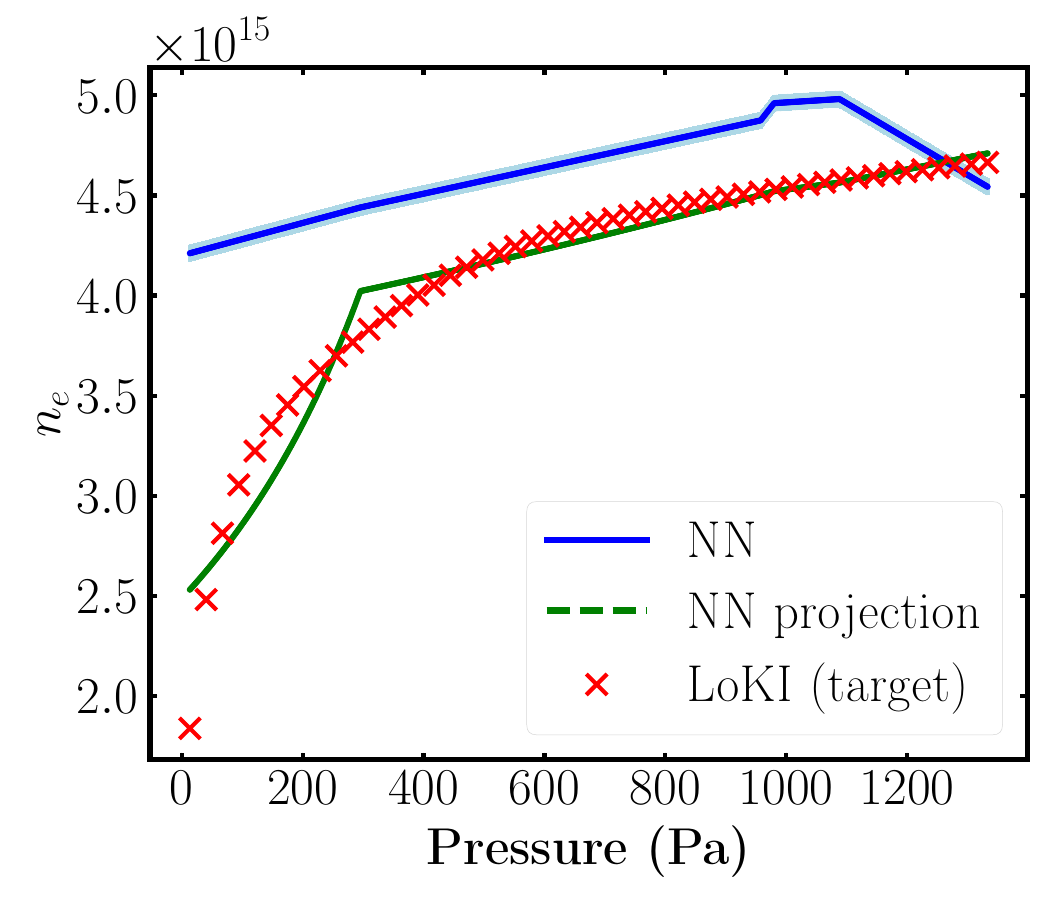}
            \put(-4,88){\textbf{(b.i)}}
        \end{overpic}
        \label{fig:4b}
    \end{subfigure}
    \begin{subfigure}[b]{0.32\textwidth}
        \centering
        \begin{overpic}[width=\textwidth]{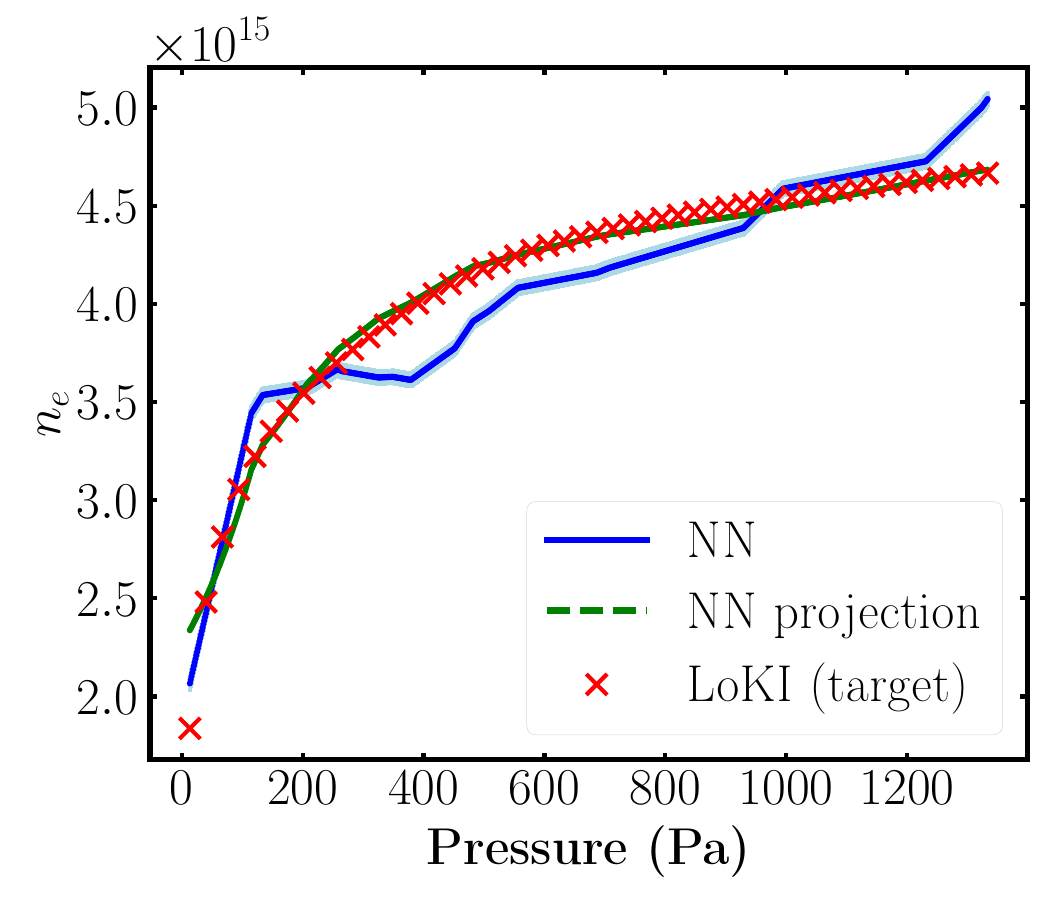}
            \put(-4,88){\textbf{(b.ii)}} 
        \end{overpic}
        \label{fig:4c}
    \end{subfigure}
    \begin{subfigure}[b]{0.32\textwidth}
        \centering
        \begin{overpic}[width=\textwidth]{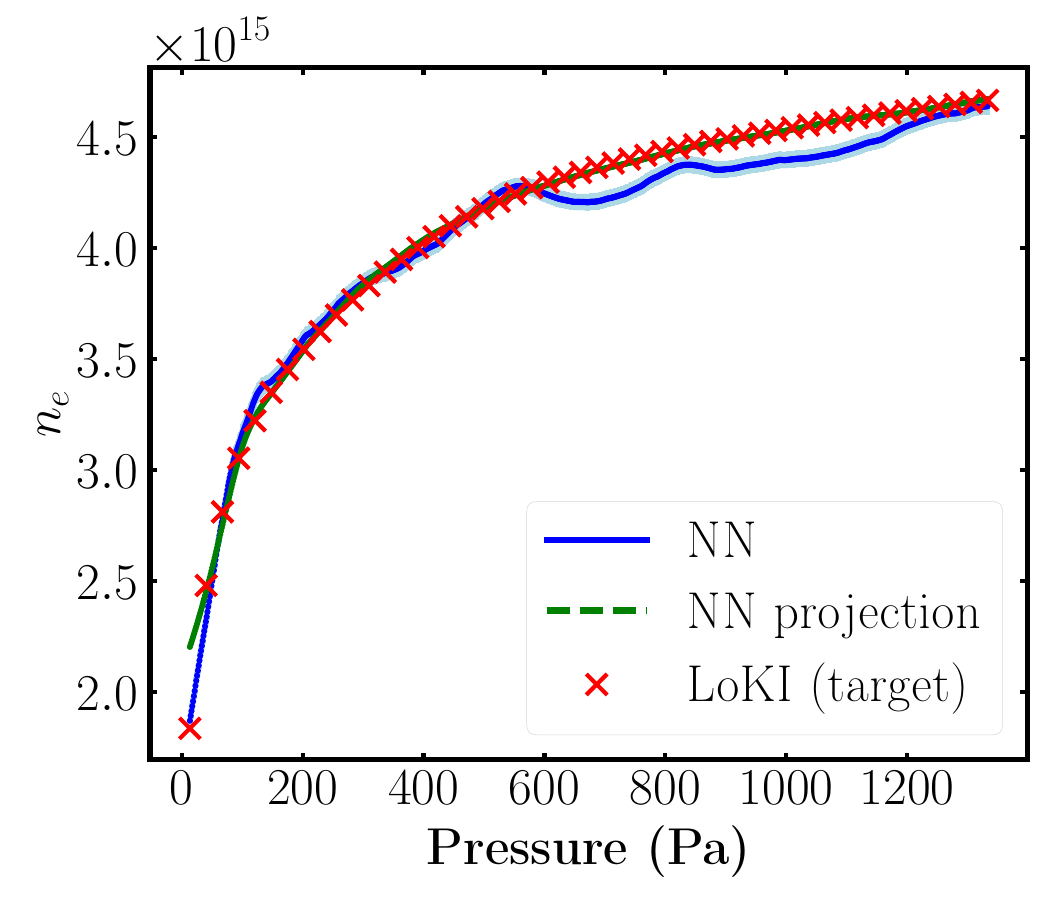}
            \put(-4,88){\textbf{(b.iii)}} 
        \end{overpic}
        \label{fig:4c}
    \end{subfigure}

    \vspace{1em}
   \begin{subfigure}[t]{0.3\textwidth}
       \centering
       \raisebox{12pt}{
           \begin{overpic}[width=\textwidth]{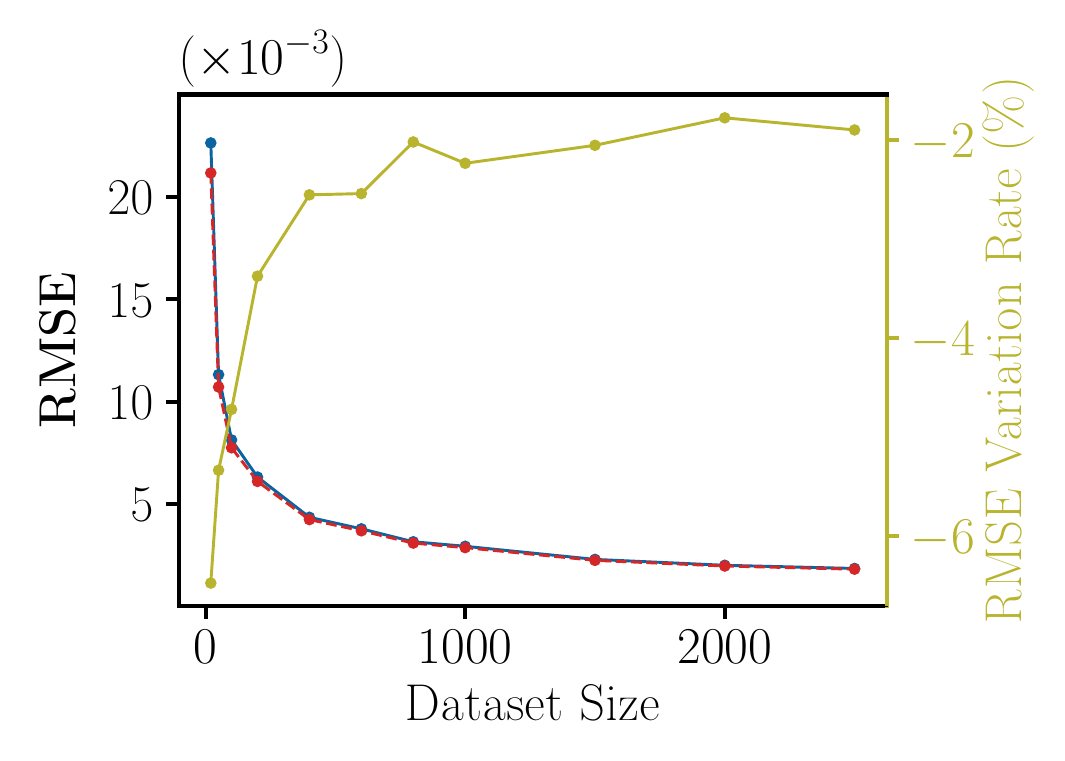}
               \put(2,80){\textbf{(c.i)}}
           \end{overpic}
       }
       \label{fig:4d}
   \end{subfigure}
   \hfill
   \begin{subfigure}[t]{0.68\textwidth}
       \centering
       \begin{overpic}[width=\textwidth]{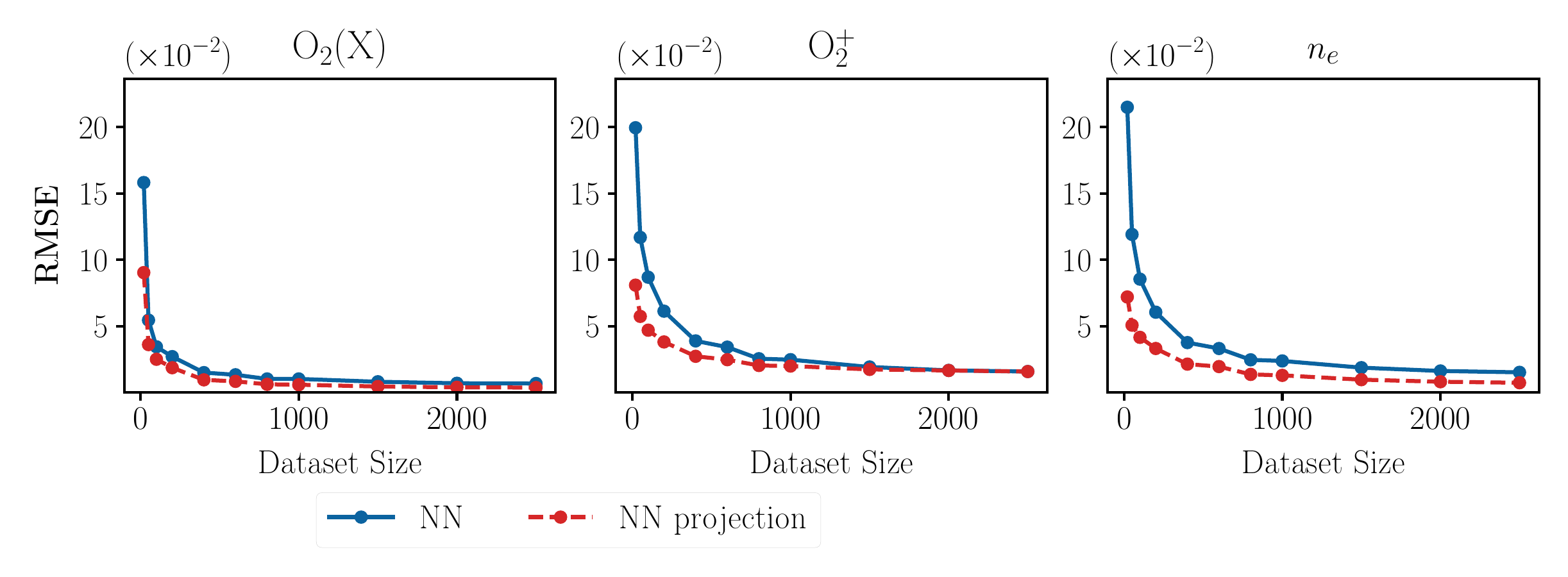}
           \put(2,40){\textbf{(c.ii)}}
       \end{overpic}
       \label{fig:4e}
   \end{subfigure}

    \begin{subfigure}[b]{0.32\textwidth}
        \centering
        \begin{overpic}[width=\textwidth]{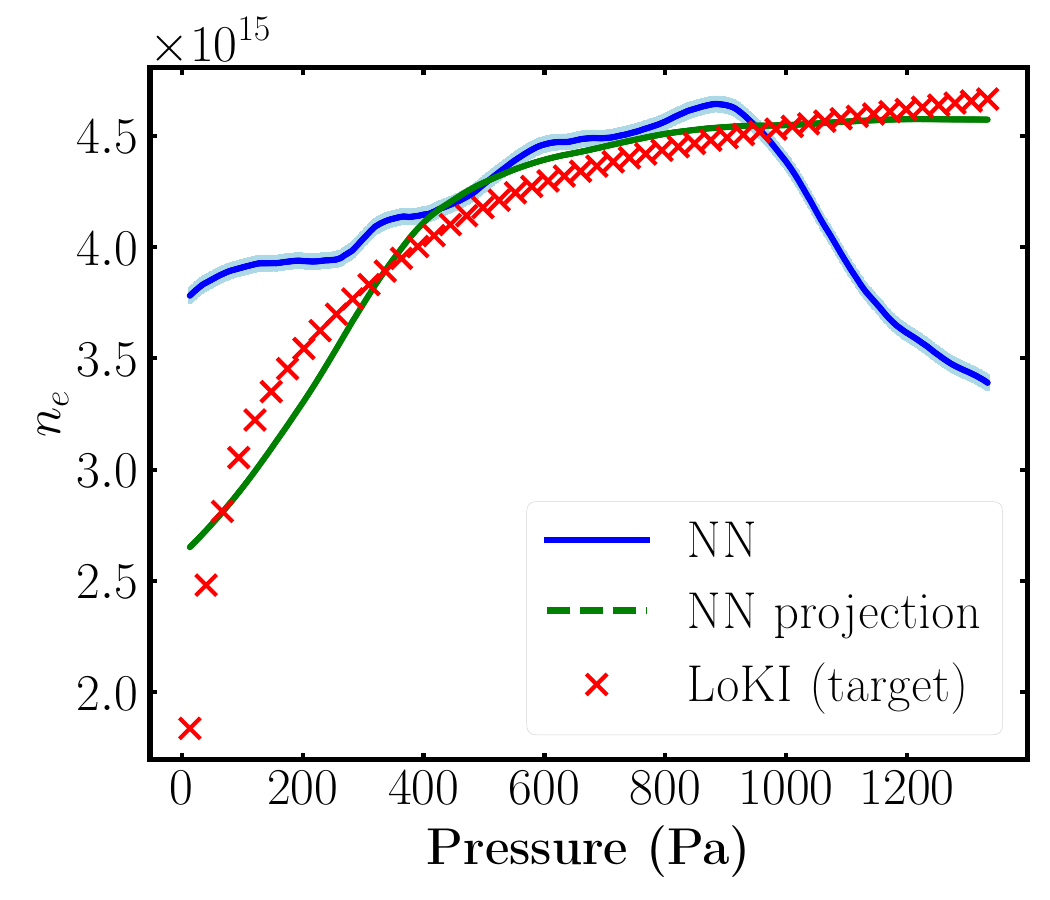}
            \put(-4,88){\textbf{(d.i)}}
        \end{overpic}
        \label{fig:4d}
    \end{subfigure}
    \begin{subfigure}[b]{0.32\textwidth}
        \centering
        \begin{overpic}[width=\textwidth]{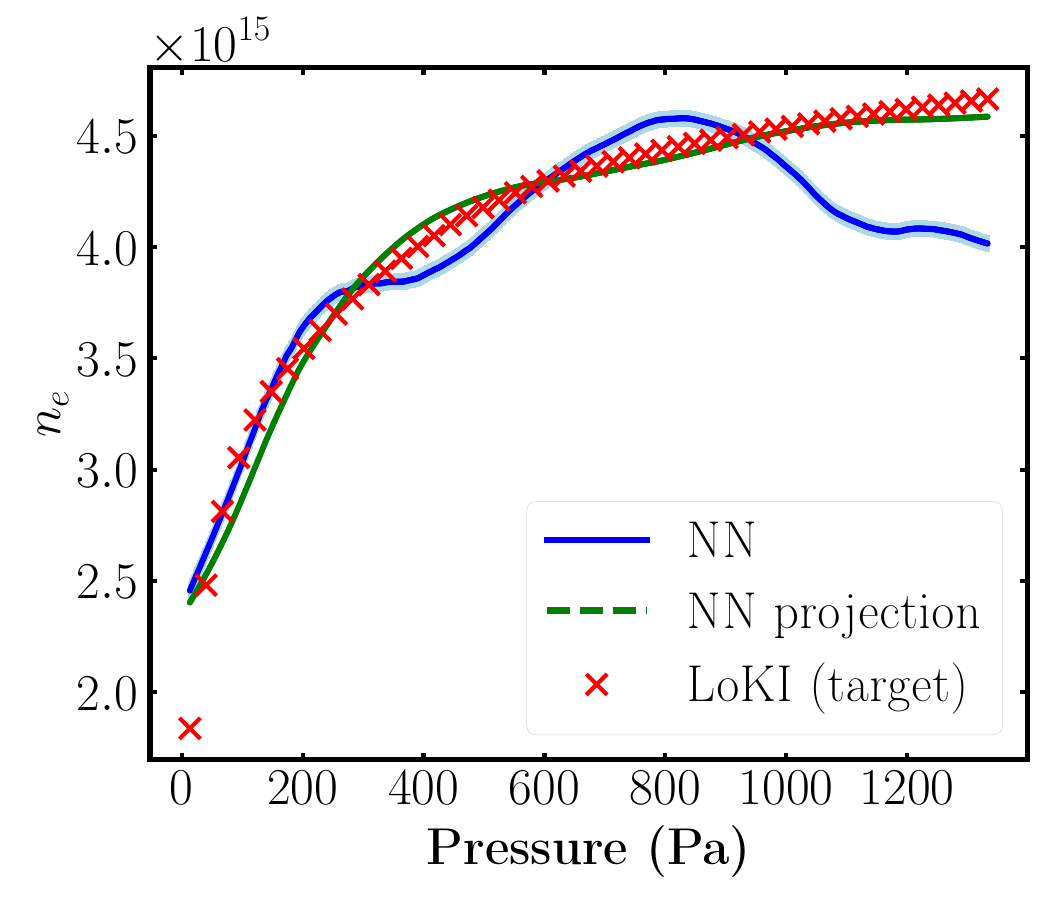}
            \put(-4,88){\textbf{(d.ii)}}
        \end{overpic}
        \label{fig:4d}
    \end{subfigure}
    \begin{subfigure}[b]{0.32\textwidth}
        \centering
        \begin{overpic}[width=\textwidth]{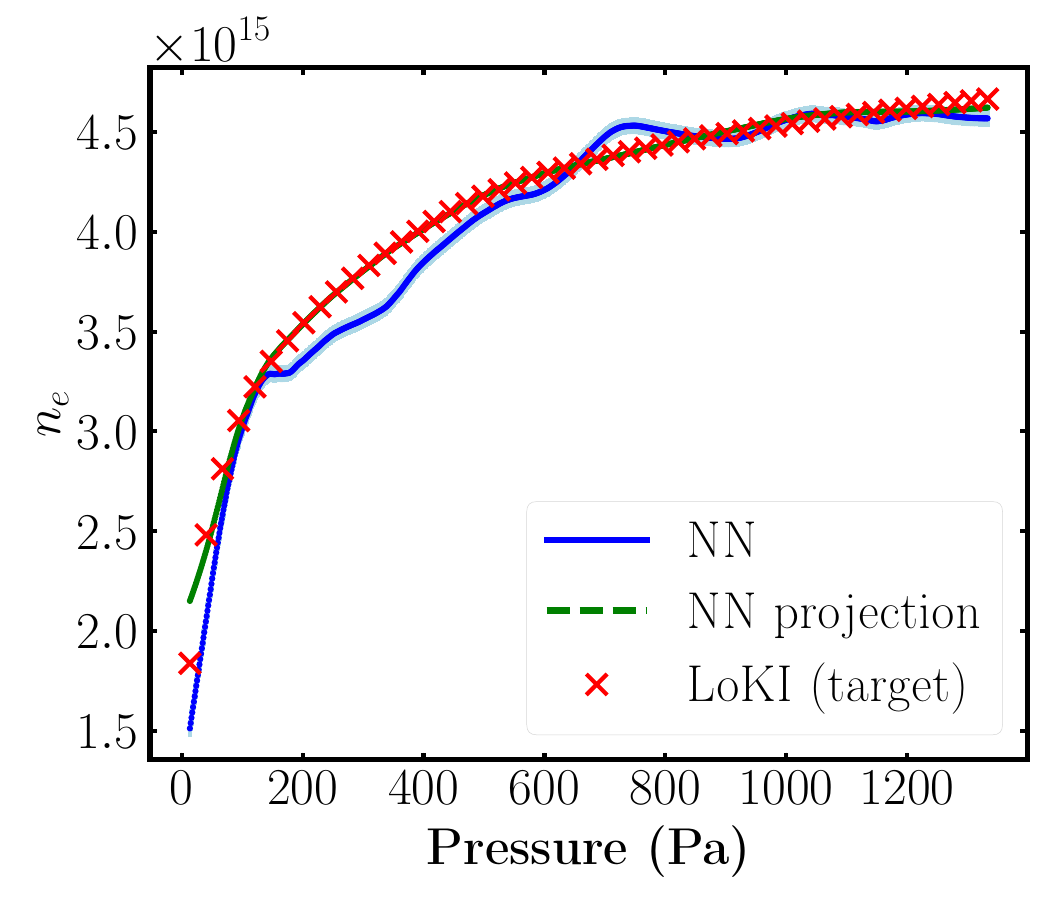}
            \put(-4,88){\textbf{(d.iii)}}
        \end{overpic}
        \label{fig:4e1}
    \end{subfigure}

    \caption{\textbf{Comparative analysis of the model performance before and after applying the projection operation to the NN outputs.} \textbf{a.} RMSE of the NN and the NN projection as a function of the model complexity (number of weights and biases in the NN architecture). \textbf{b.} NN and NN projection pressure related trends of the electron density, $n_e$, for $I=30$~mA, $R=12$~mm, and the following NN architectures: (i)  [8,8]; (ii) [26,26]; (iii) [1000,1000].  
    \textbf{c.} RMSE of the NN and the NN projection as a function of the dataset size.
    \textbf{d.} NN and NN projection pressure related trends of the electron density, $n_e$, for $I=30$~mA, $R=12$~mm, and the following dataset sizes: (i)  200; (ii) 600; (iii) 2500.}
    \label{fig:Fig_6}
\end{figure}

We now compare the errors before and after applying the projection operation to the NN predictions, considering training datasets with different sizes.
In order to guarantee a comparable analysis between the models, an architecture with 2 hidden layers and 50 neurons in each layer was used.
To mitigate randomness associated with a specific dataset sample, 20 random samples are drawn for each dataset size from a larger dataset. The NN model is then trained on each sample, and the trained models are evaluated on a test set to obtain errors for each of the 17 outputs. The results are shown in
Fig.~\ref{fig:Fig_6}c.i, where each point represents the mean error across the 20 samples for each dataset size, considering the average performance on the 17 output predictions. Furthermore, 11 different dataset sizes were analyzed, ranging from 20 to 2500 observations.
The aggregate RMSE across all 17 output quantities decreases consistently as the dataset size grows, with both methods showing similar rates of improvement. The projection method maintains a persistent advantage of approximately -2.4\% to -2.8\% in RMSE variation rate compared with the NN.

Fig.~\ref{fig:Fig_6}c.ii depicts a similar RMSE analysis as in Fig.~\ref{fig:Fig_6}c.i, but for the three most relevant individual outputs.
A similar conclusion as in the previous section can be drawn: the projected predictions outperform the non-projected counterparts, particularly for smaller dataset sizes, where the model benefits the most from the addition of physical constraints. This is further confirmed by inspection of Fig.~\ref{fig:Fig_6}d, representing the predictions of the electron density as a function of pressure, for the same discharge current and tube radius as in Fig.~\ref{fig:Fig_6}b.
With limited training data (Fig.~\ref{fig:Fig_6}d.i), the NN predicts a nonphysical behavior at higher pressures, with a sharp decrease above 800 Pa that contradicts the expected saturation pattern.
In contrast, the projection method maintains physically sound predictions across the entire pressure range, effectively compensating for the scarce training data. As expected, the accuracy of the NN progressively improves when trained on larger datasets (Fig.~\ref{fig:Fig_6}d.ii-iii), while the projection method's predictions remain more accurate and consistently aligned with physical expectations.
These results demonstrate the pivotal role of the projection method as an efficient procedure to include physical constraints into the predictions in scenarios where scarce data is available, maintaining physically meaningful predictions and substantial error reductions with limited training samples.

\newpage

\enlargethispage{25mm} % Slightly increase the page height

\subsubsection{Computation time}\label{subsec2_4_3}

\begin{figure}[htbp]
   \centering
    \begin{subfigure}[b]{0.34\textwidth}
        \centering
        \begin{overpic}[width=\textwidth]{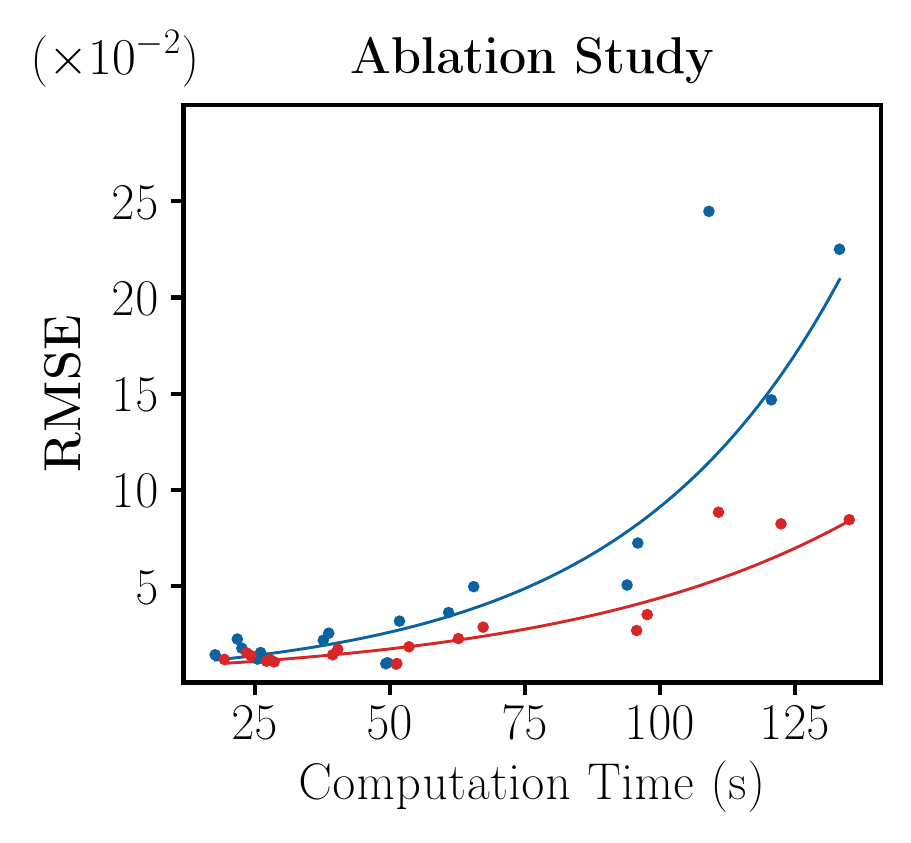}
            \put(-1,96){\textbf{(a)}}
        \end{overpic}
        \label{fig:c}
    \end{subfigure}
    \hfill
    \begin{subfigure}[b]{0.6\textwidth}
        \centering
        \begin{overpic}[width=\textwidth]{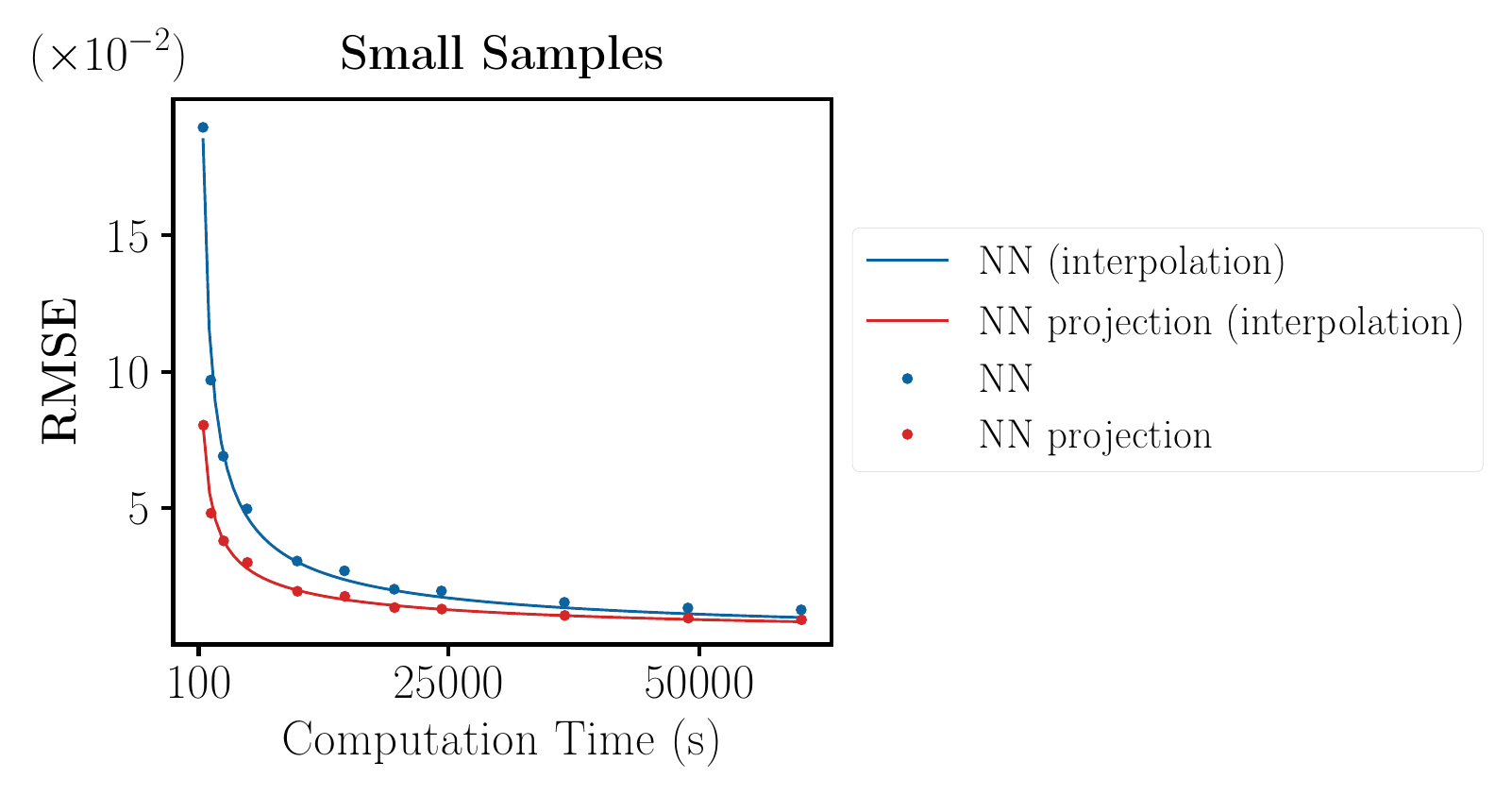}
            \put(0,54){\textbf{(b)}} 
        \end{overpic}
        \label{fig:d}
    \end{subfigure}
    \caption{\textbf{Comparative analysis of the mean RMSE of the predictions of O$_2(X)$, $\mathrm{O_2^+}$ and $n_e$ before and after applying the projection operation to the NN outputs as a function of the total computation time.} 
    \textbf{a.} Analysis for models with ranging complexity. 
    \textbf{b.} Analysis for models trained with datasets of different sizes.}
    \label{fig:Fig_7}
\end{figure}

To evaluate the computational performance of our approach, we compare the total computation time required both in the NN and the NN projection methods. Fig. \ref{fig:Fig_7}a depicts the mean RMSE of the predictions for O$_2(X)$, O$_2^+$, and $n_e$ as a function of the time required to train the models of varying complexity (described in Section \ref{subsec2_4_1}) and to evaluate their predictions in a test set. For the NN projection method, the total computation time includes the NN training and inference times, as well as the additional time required to project the outputs onto the physically consistent manifold. 
The curves correspond to an exponential fitting function to the data and are merely a guide to the eye. We observe a notable advantage of the projection method in models with longer training times. 
Moreover, on average, the projection step takes only 1.75 seconds for a dataset of 500 observations, which corresponds to a modest 4.33\% increase in the total computation time when compared with the NN model. This slight increase in computation time of the projection method comes with a reduction of the RMSE by an average of 32.08\%, with improvements, reaching up to 63.86\% in simpler architectures that require more epochs to converge. 

Fig. \ref{fig:Fig_7}b illustrates the mean RMSE of the predictions as a function of the total time required to generate data sets of different sizes (described in Section \ref{subsec2_4_2}), train the models and evaluate their predictions in a test set. 
As expected, increasing the dataset size leads to lower RMSE values for both models, but at the cost of significantly longer computation times. Clearly, the NN projection is computationally more efficient than the purely data-driven NN, despite the additional projection step.
As an example, the NN trained with 200 observations achieved a RMSE of 0.050, requiring a total computation time of 4913 seconds. 
In contrast, applying the projection method to the predictions of the NN trained with only 50 observations resulted in a comparable RMSE of 0.048, but with a significantly reduced computation time of 1338 seconds. This represents a reduction in computational cost by a factor of $\sim$3.7  while maintaining a nearly identical predictive accuracy.

\newpage

\subsection{Summary and outlook}

We introduced a flexible and effective mechanism to improve the quality of the predictions of machine learning surrogate physical models. The approach is based on projecting the model output onto the manifold defined by an arbitrary number of general physical laws, correcting the predictions after the training process, at prediction time. 
We established the proof-of-concept of our method using neural networks
and two case studies, a spring-mass system and a low-temperature reactive plasma. 
We consistently confirmed the hypothesis that our technique produces physically consistent and scientifically sound results, improves generalizability, and increases efficiency by reducing the computational and data resources necessary to build the surrogate models. 
Moreover, adding or removing physical constraints merely alters the projection manifold for the post-training predictions, without requiring any modifications to the initial model. This flexibility is a powerful tool for gaining deeper physical insight and improving the clarity of interpretation of the physical results. 

Our physics-consistent machine learning projection method can be used independently or as a complement to physics-informed neural networks. For the specific systems under study, PINN's approach of introducing physical information directly into the loss function during training was ineffective in reducing predictive errors. 
This result indicates that inclusion in the loss function of physically strong but very general laws, such as energy and charge conservation, can bring additional optimization challenges and make post-training projections more appropriate.
The difficulty is likely associated with an expansion (rather than a restriction) of accessible output parameter space related to these universal laws. 
This may occur because, when the training data already adheres to these laws, incorporating them as regularization terms adds no new meaningful information to the system. Further research is needed to clarify this issue. 
Notwithstanding, synergy between PINNs and the projection method is expected. While remaining largely unexplored, it holds the promise of further leveraging artificial intelligence to address scientific and practical engineering problems, especially in contexts of limited resources.

\section{Methods}\label{sec4}

\subsection{Data pre-processing}\label{subsec4_3}

For both the spring-mass and the low-temperature reactive plasma (LTP) systems, we used min-max scaling normalizing features to the $[-1, 1]$ interval, thereby ensuring equal contribution from all features to the loss function and mitigating magnitude disparities.
In addition, the LTP system exhibited notable skewness, particularly evident in low-pressure regimes. We quantified this skewness for each output variable and applied logarithmic transformations when the skewness exceeded a predetermined threshold. This transformation was necessary for several output parameters, specifically O($^1$D), O$^+$, $E/N$, and $T_e$.

\subsection{Model training and optimization}\label{subsec4_4}

In both case studies, fully connected feed-forward NNs were implemented in PyTorch. 
Moreover, a standard train-test-validation split of the dataset was performed to evaluate model performance, with 80\% allocated for training, 10\% for validation, and 10\% for testing.
For the relatively straightforward spring-mass system, we set a fixed training duration of 60 epochs. In turn, an early stopping criterion is used in the modeling of the highly complex LTP system.
The implemented early stopping criterion follows Prechelt's $PQ_\alpha$ early stopping method \cite{early_stopping}, an approach that aims to balance the trade-off between training time and generalization performance. The implementation handles special cases such as insufficient epochs and zero training progress scenarios, with the threshold parameter $\alpha$ controlling stopping sensitivity. 
Xavier's initial value \cite{bib_methods_xavier} is used as the initial NN parameters, which are updated using the gradient-based algorithm Adam \cite{bib_methods_adam}.

\subsubsection{Spring-mass system}\label{subsec4_4_1}
%neural network modelling
In the spring-mass system from section~\ref{subsec2_2}, the NN takes the system's current state variables $(x_1, v_1, x_2, v_2)$ and predicts the state variables in the following state.
% loss function
We use a standard mean squared error (MSE) loss function for the NN training, while for the training of the loss-based PINN an additional loss function is defined as the residuals of the energy conservation law [\textit{cf.} Fig.\ref{three_model_flowchart}b]. 
In this way, the loss terms are written as

\begin{align}
\mathcal{L}_{\text{physics}} &= \text{MSE}(E_{\text{out}}, E_{\text{in}}) \\
\mathcal{L}_{\text{data}} &= \text{MSE}(y_{\text{out}}, \hat{y})\label{LData} \\
\mathcal{L}_{\text{total}} &= (1 - \lambda_{\text{physics}}) \cdot \mathcal{L}_{\text{data}} + \lambda_{\text{physics}} \cdot \mathcal{L}_{\text{physics}} \label{LTotal}
\end{align}

\noindent where $E_{\text{out}}$ and $E_{\text{in}}$ are the system energies calculated from the output and input state variables, respectively, and $\lambda_{\text{physics}}$ is the positive weighting parameter given to the energy conservation  constraint defined in $[0,1]$.

A sensitivity analysis was performed in order to study the effect of the $\lambda_{\text{physics}}$ on the validation loss of the loss-based PINN model. Moreover, by training different PINN models with varying $\lambda_{\text{physics}}$ defined in $[0,1]$, $\mathcal{L}_{\text{total}}$, $\mathcal{L}_{\text{physics}}$ and $\mathcal{L}_{\text{data}}$ were analyzed on the validation data. The results have shown that, for this physical system, $\lambda_{\text{physics}}$ does not have a strong effect on the predictions unless the value is $\sim 1$, as in this case the weight given to the data on the optimization of the network's parameters is too low. 
The analysis in section~\ref{subsec2_2_2} is performed by selecting a value for $\lambda_{\text{physics}}$ that balances in terms of orders of magnitude the $\mathcal{L}_{\text{data}}$ and $\mathcal{L}_{\text{physics}}$ parameters in the $\mathcal{L}_{\text{total}}$, specifically a $\lambda_{\text{physics}}= 0.005$ was selected.

%
%optimization
The same NN structure was used for all examples. 
Leaky ReLU activation functions were applied between the fully connected layers of sizes [22, 98, 9].
The learning rate in the Adam algorithm \cite{bib_methods_adam} was set to $\eta= 0.0001$.
Finally, a maximum number of 60 training epochs was defined to allow the comparison between models.

\subsubsection{Low-temperature reactive plasma}\label{subsec4_4_2}
% loss function
In the low-temperature reactive plasma system studied in section~\ref{subsec2_3}, we use a standard mean squared error (MSE) loss function for the NN training, as before.
Regarding the training of the loss-based PINN, we define a loss function as the residuals of each of the three physical laws (\ref{eq_ideal_gas_law}-\ref{eq_quasi_neutrality}). The corresponding terms are written as

\begin{align}
\mathcal{L}_{\text{P}}   &= \text{MSE}(P_{\text{out}}, P_{\text{in}}) \\
\mathcal{L}_{\text{I}}   &= \text{MSE}(I_{\text{out}}, I_{\text{in}}) \\
\mathcal{L}_{\text{n}_e} &= \text{MSE}(n_{e,\text{out}}, n_{i,\text{out}}) 
\end{align}

\noindent where \( P_{\text{in}} \) and \( I_{\text{in}} \) are the input pressure and discharge current, respectively, \( P_{\text{out}} \) and \( I_{\text{out}} \) are calculated from the output predictions and input features (see Eqs.~(\ref{eq_ideal_gas_law}) and (\ref{eq_discharge_current})), \( n_{e, \text{out}} \) is the predicted electron density and $n_{i,\text{out}}$ is the predicted difference between the total positive and negative ion densities (see Eq.~(\ref{eq_quasi_neutrality})).
Consequently, the total loss function associated with the physical constraints is given by

\begin{align}
%lambda total
\lambda_{\text{physics}} &= \lambda_{\text{P}} + \lambda_{\text{I}} + \lambda_{\text{$n_e$}} \\
%physics loss
\mathcal{L}_{\text{physics}} &= \lambda_{\text{P}} \cdot \mathcal{L}_{\text{P}} + 
\lambda_{\text{I}} \cdot \mathcal{L}_{\text{I}} + 
\lambda_{\text{$n_e$}} \cdot \mathcal{L}_{\text{$n_e$}}  %\\
%loss data
%\mathcal{L}_{\text{data}} &= (1 - \lambda_{\text{physics}}) \cdot \mathcal{L}_{\text{data}} \\
\end{align}

\noindent 
where $\lambda_{\text{P}}$, $\lambda_{\text{I}}$, and $\lambda_{\text{$n_e$}}$ are positive weighting parameters given to each physical constraint defined in $\lambda_{\text{physics}} \in [0,1]$, 
while $\mathcal{L}_{\text{data}}$ and $\mathcal{L}_{\text{total}}$ are given as before, by Eqs.~(\ref{LData}) and (\ref{LTotal}).
Given that the results of the analysis of the previous physical system revealed no clear correlation between the $\lambda_{\text{physics}}$ parameter and the performance of the PINN model in the validation set, a $\lambda_{\text{physics}}$ was selected to balance in terms of orders of magnitude the values of $\mathcal{L}_{\text{data}}$ and $\mathcal{L}_{\text{physics}}$ in $\mathcal{L}_{\text{total}}$. 
Moreover, the same weight was given to each of the three laws, being $\lambda_{\text{physics}}= 0.015$.

%optimization
The same NN structure, with Leaky ReLU activation functions applied between 2 fully connected layers of sizes [50, 50], is used for all examples, except for the analysis in section \ref{subsec2_4_1}. 
%Xavier's initial value \cite{bib_methods_xavier} was used as the initial NN parameters.
%The NN parameters are updated using the gradient-based algorithm Adam \cite{bib_methods_adam} with 
The learning rate is updated dynamically to ensure a more efficient training process.
In this way, by monitoring the convergence of the validation loss during the training process, the learning rate is reduced by a factor of 10 when the validation loss reaches a plateau.

Finally, to reduce the dependence on the seed selection and to obtain statistically significant results, a bootstrapped model was trained.
In particular, in the results in section \ref{subsubsec2_3_3}, $N=30$ different NN models initialized with different seeds are trained.
The final model prediction is given by the mean of the individual predictions of the $N$ models, and a quantification of the uncertainty of the prediction is obtained by computing the corresponding standard deviation.
%$\frac{\sigma_{predictions}}{\sqrt{N}}$.

\newpage

\subsection{The projection method}\label{subsec4_5}

Our projection method can be formalized as follows.
Consider a ML parametric model $y=f(x; \Theta)$, where $x$ is the input, $y$ the output, and $\Theta$ is the model parameter vector.
Given a set $\mathcal{D} = \{(x_k, y_k)\}$ of datapoints for training and assuming a loss function $\mathcal{L}(y_1, y_2)$, training the model amounts to solve the optimization problem 

\begin{equation}
    \underset{\Theta}{\text{minimize}} \sum_{(x,y) \in \mathcal{D}} \mathcal{L}(y, f(x;\Theta))\ .
\label{eq_ml_loss_minimization}
\end{equation}
Additionally, the requirement that a set of physical laws relating both $x$ and $y$ must be satisfied is considered.
This requirement can be expressed as a vector valued constraint function $g(x,y)$ that is zero if, and only if, those physical laws are satisfied, \textit{i.e.}, $g(x,y) = 0$.
Even though the input dataset $\mathcal{D}$ might satisfy this constraint, there is not \textit{a priori} any guarantee that the ML model $f$ will output values that satisfy these laws.

One common approach to add physical information to a ML model is to include into the loss function $\mathcal{L}$ a regularization term that penalizes violations of the physical constraints.
However, once the model is trained, there is still no guarantee that the outputs for unseen inputs do satisfy those constraints.
Moreover, if the physical laws are too general, they may fail to guide the NN and improve its predictions, as shown in this work.
Here we follow an alternative procedure, that explores the idea of projecting the output $y$ of the model onto the manifold defined by the constraint $g(x,y)=0$.

The projection operation is formulated as the constraint optimization problem defined by equation (\ref{eq_projection_constraint_problem}) and
was implemented using CasADI opti stack \cite{casadi_ref}.
Specifically, the nonlinear programming solver IPOPT was used to solve the optimization problem with tolerances set to \(10^{-8}\) for the LTP system and \(10^{-3}\) for the spring-mass system.
In addition, all along the paper we performed the projection with the identity matrix, $W=\mathds{1}$. Strategies to optimizing the projection with a different weighting matrix can be explored and are left for future work.

\section{Data availability}\label{sec5}

The datasets generated in this study are available on Github [\url{https://github.com/matildevalente/physics_consistent_machine_learning}].

\section{Code availability}\label{sec6}

The code that supports the findings in this study is available on Github [\url{https://github.com/matildevalente/physics_consistent_machine_learning}].

\section*{Acknowledgements}\label{sec8}

This work was supported by the Portuguese FCT - Fundação para a Ciência e a Tecnologia, under funding to LARSyS (DOI: 10.54499/LA/P/0083/2020, 10.54499/UIDP/50009/2020, and 10.54499/UIDB/50009/2020), to IPFN (DOI: 10.54499/UIDB/50010/2020, 10.54499/UIDP/50010/2020, and 10.54499/LA/P/0061/2020), and to project
PTDC/FIS-PLA/1616/2021 (DOI: 10.54499/PTDC/FISPLA/1616/2021). 
We would like to thank Dr. Alcides Fonseca for the very helpful comments and suggestions to improve this paper.

%\section*{Author contributions}

%Following this structure:
%https://www.elsevier.com/researcher/author/policies-and-guidelines/credit-author-statement

%MV: Software, Formal analysis, Investigation, Original draft, visualization; TCD: Methodology, Software, Investigation, Review \& editing; RV: Conceptualization, Methodology, Resources, Review \& editing, Supervision, Funding acquisition; VG: Methodology, Review \& editing, Supervision, Funding acquisition.

%Nature seems more relaxed...


\begin{thebibliography}{99}\label{sec7}

\bibitem{Nyshadham2019}
Nyshadham, C., Rupp, M., Bekker, B. et al. Machine-learned multi-system surrogate models for materials prediction. \textit{npj Comput Mater} \textbf{5}, 51 (2019). https://doi.org/10.1038/s41524-019-0189-9.

\bibitem{Schmidt2019}
Schmidt, J., Marques, M., Botti, S. \& Marques, M. Recent advances and applications of machine learning in solid-state materials science. \textit{npj Comput Mater} \textbf{5} (2019). https://doi.org/10.1038/s41524-019-0221-0.

\bibitem{Jiang2021}
Jiang, J., Chen, M. \& Fan, J.A. Deep neural networks for the evaluation and design of photonic devices. \textit{Nat Rev Mater} \textbf{6}, 679–700 (2021). https://doi.org/10.1038/s41578-020-00260-1.

\bibitem{Pache2022}
Pache, R. \& Rung, T. Data-driven surrogate modeling of aerodynamic forces on the superstructure of container vessels.
\textit{Eng Appl Comput Fluid Mech} \textbf{16}, 746–763 (2022). https://doi.org/10.1080/19942060.2022.2044383.

\bibitem{Dakane2024}
Dhakane, V. \& Yada, A. Computational Fluid Dynamics–Deep Neural Network (CFD-DNN) Surrogate Model with Graphical User Interface (GUI) for Predicting Hydrodynamic Parameters in Three-Phase Bubble Column Reactors. 
\textit{Ind Eng Chem Res} \textbf{63}, 11670–11685 (2024). https://doi.org/10.1021/acs.iecr.4c00669.

\bibitem{Zhang2025}
Zhang, W., Zhang, C., Zhao, Y. et al. Convolutional neural networks-based surrogate model for fast computational fluid dynamics simulations of indoor airflow distribution. \textit{Energy Build} \textbf{326}, 115020 (2025). https://doi.org/10.1016/j.enbuild.2024.115020.

\bibitem{Bonzanini2023}
Bonzanini, A.D., Shao, K., Graves, D.B. et al. Foundations of machine learning for low-temperature plasmas: methods and case studies. \textit{Plasma Sources Sci Technol} \textbf{32}, 024003 (2023). https://doi.org/10.1088/1361-6595/acb28c.

\bibitem{Liu2024}
Liu, P., Wu, Q., Ren, X., Wang, Y. \& Ni, D. A deep-learning-based surrogate modeling method with application to plasma processing. \textit{Chem Eng Res Des} \textbf{211}, 299-317 (2024). https://doi.org/10.1016/j.cherd.2024.09.031.

\bibitem{Zhao2024}
Zhao, Y., Chen, W., Miao, Z., Yang, P. \& Zhou, X. Deep learning-assisted magnetized inductively coupled plasma discharge modeling. \textit{Plasma Sources Sci Technol} \textbf{33}, 125013 (2024). https://doi.org/10.1088/1361-6595/ad98bf.

\bibitem{Diaw2021}
Diaw, A., McKerns, M., Sagert, I. et al. Efficient learning of accurate surrogates for simulations of complex systems. \textit{Nat Mach Intell} \textbf{6}, 568–577 (2024). https://doi.org/10.1038/s42256-024-00839-1.

\bibitem{Raissi2019} 
Raissi, M., Perdikaris, P. \& Karniadakis, G.E. Physics-informed neural networks: A deep learning framework for solving forward and inverse problems involving nonlinear partial differential equations, \textit{J Comp Phys} \textbf{378}, 686--707, 2019.
https://doi.org/10.1016/j.jcp.2018.10.045.

\bibitem{intro_pinns_flows}
Wong, H.S., Chan, W.X., Li, B.H. et al. Strategies for multi-case physics-informed neural networks for tube flows: a study using 2D flow scenarios. \textit{Sci Rep} \textbf{14}, 11577 (2024). https://doi.org/10.1038/s41598-024-62117-9.

\bibitem{intro_pinns_materials}
Pun, G.P.P., Batra, R., Ramprasad, R. et al. Physically informed artificial neural networks for atomistic modeling of materials. \textit{Nat Commun} \textbf{10}, 2339 (2019). https://doi.org/10.1038/s41467-019-10343-5.

\bibitem{Huang2023}
Huang, B. \& Wang, J. Applications of Physics-Informed Neural Networks in Power Systems - A Review. \textit{IEEE Trans Power Sys} \textbf{38}, 572--588 (2023). https://doi.org/10.1109/TPWRS.2022.3162473.

\bibitem{Moschou2023}
Moschou, S.P.,  Hicks, E.,  Parekh, R.Y. et al. Physics-informed neural networks for modeling astrophysical shocks. \textit{Mach Learn: Sci Technol} \textbf{4}, 035032 (2023). https://doi.org/10.1088/2632-2153/acf116.

\bibitem{intro_pinns_fusion}
Seo, J., Kim, I.H., Nam, H.
Leveraging physics-informed neural computing for transport simulations of nuclear fusion plasmas.
\textit{Nucl Eng Technol} \textbf{56},
5396--5404 (2024). https://doi.org/10.1016/j.net.2024.07.048.

\bibitem{Kashinath2021}
Kashinath, K., Mustafa, M., Albert, A. et al. Physics-informed machine learning: case studies for weather and climate modelling. \textit{Phil Trans R Soc A} \textbf{379}, 20200093 (2021). https://doi.org/10.1098/rsta.2020.0093.

\bibitem{Karniadakis2021}
Karniadakis, G.E., Kevrekidis, I.G., Lu, L. et al. Physics-informed machine learning. \textit{Nat Rev Phys} \textbf{3}, 422–440 (2021). https://doi.org/10.1038/s42254-021-00314-5.

\bibitem{Cuomo2022}
Cuomo, S., Di Cola, V.S., Giampaolo, F. et al. Scientific Machine Learning Through Physics–Informed Neural Networks: Where we are and What’s Next. \textit{J Sci Comput} \textbf{92}, 88 (2022). https://doi.org/10.1007/s10915-022-01939-z.

\bibitem{Seo2024}
Seo, J. Solving real-world optimization tasks using physics-informed neural computing. \textit{Sci Rep} \textbf{14}, 202 (2024). https://doi.org/10.1038/s41598-023-49977-3.

\bibitem{Bolton2019}
Bolton, T., \& Zanna, L. Applications of deep learning to ocean data inference and subgrid parameterization. \textit{J Adv Model Earth Syst} \textbf{11}, 376–399 (2019).  https://doi.org/10.1029/2018MS001472

\bibitem{Jin2020}
Jin, P., Zhang, Z., Zhu, A. et al. SympNets: Intrinsic structure-preserving symplectic networks for identifying Hamiltonian systems. \textit{Neural Netw} \textbf{132}, 166-179 (2020).
https://doi.org/10.1016/j.neunet.2020.08.017.

\bibitem{Beucler2021}
Beucler, T., Pritchard, M., Rasp S. et al.
Enforcing Analytic Constraints in Neural Networks Emulating Physical Systems.
\textit{Phys Rev Lett} \textbf{126}, 098302 (2021).
https://doi.org/10.1103/PhysRevLett.126.098302

\bibitem{Tong2021}
Tong, Y., Xiong, S., He, X. et al.
Symplectic neural networks in Taylor series form for Hamiltonian systems.
\textit{J Comp Phys} \textbf{437}, 110325 (2021). https://doi.org/10.1016/j.jcp.2021.110325.

% Referencia artigo Tiago com o reaction mechanism
\bibitem{bib_intro_13}
Dias, T.C., Fromentin, C., Alves, L.L. et al. A reaction mechanism for oxygen plasmas. 
\textit{Plasma Sources Sci Technol} \textbf{32}, 084003 (2023). https://doi.org/10.1088/1361-6595/aceaa4.

% Referencia LoKI
\bibitem{bib_intro_3}
A Tejero-del-Caz, V Guerra, D Gonçalves, M Lino da Silva, L Marques, N Pinhão, C D Pintassilgo, and L L Alves,
The LisbOn KInetics Boltzmann solver,
\textit{Plasma Sources Sci. Technolo.} \textbf{28}, pages 043001 (2019). https://dx.doi.org/10.1088/1361-6595/ab0537.

\bibitem{early_stopping}
Prechelt, L. 
Early Stopping - But When?. In: Orr, G.B., Müller, KR. (eds) Neural Networks: Tricks of the Trade.,
\textit{Lect. Notes Comput. Sci.} \textbf{1524}. Springer, Berlin, Heidelberg.

% Xavier algorithm
\bibitem{bib_methods_xavier}
Glorot, X., \& Bengio, Y. Understanding the difficulty of training deep feedforward neural networks. In Y. W. Teh \& M. Titterington (Eds.), \textit{Proceedings of the Thirteenth International Conference on Artificial Intelligence and Statistics}, \textit{PMLR} \textbf{9}, 249–256 (2010). https://proceedings.mlr.press/v9/glorot10a.html

% Adam algorithm
\bibitem{bib_methods_adam}
Kingma, D. P., \& Ba, J. Adam: A method for stochastic optimization. arXiv (2017). https://arxiv.org/abs/1412.6980

\bibitem{casadi_ref}
Joel A E Andersson and Joris Gillis and Greg Horn and James B Rawlings and Moritz Diehl, \textit{CasADi -- {A} software framework for nonlinear optimization and optimal control}, Mathematical Programming Computation, 2018


\end{thebibliography}
\end{document}